%% file: main.tex
\documentclass[11pt]{article}

\usepackage[final]{acl}

\hypersetup{
    colorlinks=true,
    linkcolor=magenta,
    citecolor=magenta,
    urlcolor=magenta,
}
\usepackage{times}
\usepackage{latexsym}
\usepackage{pifont}   
\usepackage{enumitem}
\usepackage{amsmath}
\usepackage{amssymb}

\usepackage{booktabs} 
\usepackage{booktabs}
\usepackage{bbm} 
\usepackage{tabularx}
\usepackage{float}
\usepackage{xcolor}
\newcommand{\ua}[1]{{\color{red}$\uparrow$\,#1}}    
\newcommand{\da}[1]{{\color{blue}$\downarrow$\,#1}}  

\usepackage[T1]{fontenc}

\usepackage[utf8]{inputenc}

\usepackage{microtype}

\usepackage{inconsolata}

\usepackage{graphicx}
\usepackage{stfloats} 

\title{Narrative Flattening: How Post-Training Compresses Thematic, Affective, and Stylistic Variation in LLM Fiction}

\author{
  \textbf{Zehan Li\textsuperscript{1,}\textsuperscript{\ding{41}}} \quad
  \textbf{Yutong Zhu\textsuperscript{1}} \quad
  \textbf{Siyang Wu\textsuperscript{1}} \quad
  \textbf{Honglin Bao\textsuperscript{1,}\textsuperscript{\ding{41}}} \quad
  \textbf{James A.~Evans\textsuperscript{1}}
\\
\\
  \textsuperscript{1}Knowledge Lab, University of Chicago
\\
    \textsuperscript{\ding{41}} Correspondence:
    H.B. \href{mailto:honglinbao@uchicago.edu}{\texttt{honglinbao@uchicago.edu}};\\
    Z.L. \href{mailto:zehan}{\texttt{zehan@uchicago.edu}}
}

\begin{document}
\maketitle

\input{Sections/00-abstract}

\input{Sections/01-intro}

\input{Sections/02-related-work}

\input{Sections/03-Experimental-Design}

\input{Sections/04-Results}

\input{Sections/06-Discussion}

\input{Sections/07-Conclusion}

\input{Sections/08-Limitations}

\input{Sections/Ethical_Statement}

\bibliography{custom}

\input{Sections/Appendix}

\end{document}

%% file: Sections/00-abstract.tex
\begin{abstract}
Large language models produce fluent fiction, yet their creative output is widely seen as flat. 
We ask where this quality originates in the training and whether it affects different domains of human fiction equally.
We construct a matched story-continuation paradigm across StoryStar (public-platform), TMAS (prompt-guided), and \textit{The New Yorker} (professional literary)—and compare continuations from four OLMo 32B checkpoints (Base, SFT, DPO, RLVR) against matched human text.
Because these checkpoints share architecture, scale, tokenizer, and pretraining, the design isolates the post-training effect. 
We measure each continuation along three sentence-level dimensions: thematic motion, affective prevalence, and linguistic diversity. 
Across all three, post-training compresses dynamic variation: thematic transitions become more uniform, high-intensity emotions give way to neutrality, and stylistic diversity across stories shrinks. We term this progressive loss \emph{narrative flattening}. 
The effect is directionally stable across story domains but gap size depends on the human baseline: professional literary fiction is compressed most, while public-platform and prompt-guided stories show smaller gaps, consistent with their human baselines sitting closer to the model's default rhythm.
Post-trained endpoints converge across domains, suggesting alignment produces a continuation regime largely insensitive to the source domain's narrative texture.
\end{abstract}

%% file: Sections/01-intro.tex
\section{Introduction}
\label{sec:intro}

\begin{figure}[t]
    \includegraphics[width=\columnwidth]{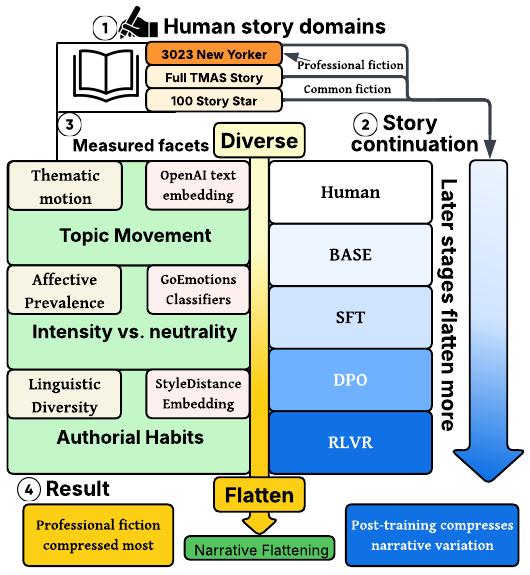}
    \caption{\textbf{Matched-continuation pipeline for measuring  post-training effects on creative writing.} We collect short stories from three human writing domains, truncate each at four  prefix lengths (40/60/80/90\%), and complete each prefix with four OLMo-32B checkpoints (Base, SFT, DPO, RLVR). Continuations are analyzed along three narrative facets: thematic motion  (sentence embeddings), affective prevalence (emotion classifier),  and linguistic diversity (style embeddings). We measure how each post-training stage reshapes the dynamic structure of story continuations---and how the reshaping depends on the source domain.}
    \label{fig:pipeline}
    \vspace{-4mm}
\end{figure}

Large language models are increasingly deployed as creative-writing assistants. Yet, a recurring complaint has emerged: LLM-generated fiction tends toward clich\'{e}, formulaic structure, stylistic monotony, and cross-piece homogeneity \citep{chakrabarty2024art, chakrabarty2026can}, and that human-AI co-writing reduces diversity across outputs \citep{doshi2024generative}.
These critiques remain at the level of lexical diversity scores, preference ratings, and whole-text quality judgments, none of which decomposes \textit{how} a story becomes flat into measurable narrative dimensions.
Missing is a mechanical account: \emph{which properties} of narrative does model generation compress?
Is the perceived flatness a matter of thematic movement, emotional dynamics, stylistic range---or all of these at once?
Without decomposing the complaint into measurable components, `AI writing feels flat' remains an observation, not a diagnosis.

Modern assistants pass through supervised fine-tuning (SFT), preference optimization (DPO), and reinforcement learning (RLVR)---stages that optimize for coherence, helpfulness, and human preference \citep{ouyang2022training, lambert2024tulu3}, and that are known to narrow output diversity even on creative tasks \citep{kirk2024understanding, padmakumar2024does, omahony2024attributing, murthy2025one}. Yet few studies trace how each stage reshapes the texture of story generation, nor whether this convergence compresses dimensions that distinguish literary from generic prose.

Human creative writing is not a single baseline. Professionally edited literary fiction, public-platform stories, and prompt-guided narratives differ systematically in topic, perceived quality, and in how they modulate themes, emotional intensity, and stylistic density \citep{biber2009register, underwood2016longue}.
If post-training converges to a fixed continuation regime, the gap should be \emph{domain-sensitive}: the farther a human writing domain lies from that regime, the more its narrative texture will be compressed.
A single-baseline evaluation risks overstating or obscuring the effect; a cross-domain comparison is  needed.

We call this pattern \emph{narrative flattening}.
A story traces a trajectory through thematic, affective, and linguistic space, and what distinguishes human fiction is not smoothness but structured variation---when to dwell on a motif, when to shift abruptly, when to heighten or suppress emotional pressure.
We define \textit{narrative flattening} as the measurable compression of this variation—regularized movement, muted affect, and reduced sensitivity to the source domain.

This paper asks two questions.
First, how does each stage of the post-training pipeline---Base, SFT, DPO, RLVR---reshape the dynamic structure of story continuations?
Second, does the magnitude of this reshaping depend on the human writing domain that the model is asked to continue?

To answer both, we construct a matched continuation paradigm spanning three distinct human writing domains in the creative-writing landscape: \textsc{StoryStar}, a public fiction platform with a low editorial barrier \citep{storystar}; \textsc{Tell Me A Story}, a corpus of stories composed under explicit creative-writing prompts \citep{huot2024agents}; and \textit{The New Yorker}, a collection of professionally edited literary fiction published between 1945 and 2019 \citep{shaalan2022view}.
We truncate each story at controlled prefix lengths and collect continuations from four checkpoints along the OLMo 32B instruction path \citep{olmo2025olmo2}.
Because these checkpoints share architecture, scale, tokenizer, and pretraining, the design lets us trace how successive post-training stages reshape narrative distributions within a single model lineage, while avoiding cross-model confounds in architecture and pretraining. We treat the resulting evidence as stage-wise and geometric rather than mechanistic: the design identifies where the post-trained endpoints move relative to human baselines, but not which specific data mixtures or reward signals cause the movement.
We measure each continuation along three facets: thematic motion, affective prevalence, and linguistic diversity.

We find that each successive post-training stage narrows narrative variation across every facet we measure.
In \emph{thematic motion}, models traverse topics in more uniform steps, with per-story topic-jump variation falling below the human level by RLVR, dampening the alternation between dwelling and sharp shifts that characterizes human stories.
In \emph{affective prevalence}, high-arousal states such as surprise--curiosity and conflict are suppressed while neutral affect rises.
In \emph{linguistic diversity}, outputs collapse into a narrow stylistic attractor: closer to common-fiction baselines, but farther from professional literary style.

This pattern is \emph{directionally stable} across three story domains, but its magnitude depends on the human baseline.
Professional literary fiction is compressed most: its human distribution lies farthest from the post-trained model's default rhythm, so the narrowing is most visible.
Public-platform and prompt-guided stories show smaller gaps---not because the model writes them better, but because these baselines already sit closer to where post-training converges.
Strikingly, post-trained endpoints \emph{converge} across corpora, with cross-domain style divergence dropping by over $90\%$ from the human level to RLVR.
This convergence suggests that post-training pushes all stories toward a single continuation pattern rather than adapting to each domain's complexity.

Together, these results reveal that post-training does not simply produce better fiction.
It installs a fixed writing regime---coherent, readable, and conventionally well-formed, but flatter than any of the human domains it is asked to continue, with the largest gap appearing for the most stylistically and affectively varied human corpus.

Our contributions are threefold:
\begin{itemize}[noitemsep, topsep=2pt, leftmargin=*]
    \item We introduce a matched continuation paradigm that traces post-training effects across human writing domains differing in editorial gatekeeping and task framing.
    \item We operationalize `AI flatness' as narrative flattening: measurable compression along thematic, affective, and linguistic dimensions.
    \item We show that post-training compresses narrative texture across three domains, and most for professional literary fiction, converging post-trained endpoints toward a single mundane regime. \looseness=-1
\end{itemize}

%% file: Sections/02-related-work.tex
\section{Related Work}
\label{sec:related-work}

\subsection{LLM Creative Writing and Its Evaluation}
\label{sec:rw-creative-writing}

LLMs are increasingly used for creative writing, supporting story generation, co-writing, and screenplay assistance \citep{ippolito2022creativewritingaipoweredwriting, Lee_2022, yuan2022wordcraft}.
Most evaluations rely on holistic judgments of quality, novelty, or satisfaction. 
This framing has surfaced promise and limits: model outputs can be fluent and useful, yet also clich\'{e}d, formulaic, and stylistically monotonous \citep{chakrabarty2024art, chakrabarty2025salvaged, chakrabarty2026can, sui2026llmuncertainty, marco2024pronvsprompt}. 
Beyond individual quality, LLM assistance can homogenize collective output: \citet{doshi2024generative} find that AI co-writing improves individual stories while reducing diversity across stories; \citet{padmakumar2024does} show that instruction-tuned models reduce lexical and content diversity; and \citet{xu2025echoes} find that different aligned LLMs independently recycle the same narrative moves. 
\citet{russell2026storyscope} push beyond surface statistics, showing that AI stories over-explain themes, favor tidy single-track plots, and cluster in a narrow region of discourse-level narrative space that human stories span more broadly.

These studies identify the symptom but not its mechanism: holistic ratings and endpoint comparisons cannot distinguish whether flatness reflects predictable topic shifting, smoothed emotional pressure, converging styles, or all three. We address this gap by decomposing narrative flattening along these dimensions.

\subsection{Post-Training and Homogenization in Open-Ended Generation}
\label{sec:rw-posttraining}
Modern assistants undergo post-training through supervised fine-tuning (SFT), preference optimization (DPO), and reinforcement learning (RL), which improves instruction following but reshapes output distributions \citep{ouyang2022training, rafailov2024directpreferenceoptimizationlanguage, lambert2024tulu3}. 
\citet{kirk2024understanding} show that RLHF reduces output diversity relative to SFT; \citet{omahony2024attributing} connect instruction and reward-based tuning to mode collapse; and \citet{murthy2025one} find that aligned models display less conceptual diversity than their base counterparts. 
\citet{west2025base} sharpen the contrast: base checkpoints outperform aligned versions on tasks demanding unpredictable output, creative writing among them, and \citet{zhang2025noveltybench} find that leading aligned systems produce substantially less variation than human writers.

Recent work has begun to explain and mitigate this narrowing. 
On the diagnostic side, \citet{zhang2025verbalized} argue that annotators' systematic preference for familiar text is a pervasive data-level driver of mode collapse, and \citet{hamilton2026elias} trace the recurring names, settings, and professions in LLM stories to preference data rather than pretraining corpora. 
On the intervention side, \citet{lanchantin2025diverse} construct preference pairs favoring rare but high-quality responses, and \citet{chung2025modifying} add deviation terms to DPO objectives to retain variation. 
Both lines operate at the lexical level---recycled tokens, embedding distances---without specifying which narrative properties are worth preserving; our facets are intended to supply exactly those targets.

Creative writing is a particularly sensitive setting: thematic shifts, affective volatility, and stylistic breadth are part of the task, not noise to eliminate. 
Yet studies of alignment side effects rarely examine creative generation, and those that do typically compare a base model to a single aligned endpoint without tracing the contribution of each stage. 
We follow one model lineage across four checkpoints---Base, SFT, DPO, and RLVR---tracing stage-wise shifts in narrative texture within a single model family.

\subsection{Computational Accounts of Narrative Texture}
\label{sec:rw-narrative-texture}

Narrative theory treats stories as organized trajectories, not unordered sentences. Classic accounts emphasize how suspense, curiosity, and surprise arise from temporal ordering and revelation of events \citep{sternberg1978expositional, BREWER1982473}. 
Computational work operationalizes this at scale: \citet{reagan2016emotional} extract emotional arcs using sentiment trajectories; \citet{ouyang2015modeling} model reportable events as turning points; and \citet{piper2023modeling} model narrative revelation using information-theoretic methods. 
\citet{piper2021narrative} argues that computational narrative understanding must account for event ordering, temporality, and discourse structure. 
A parallel tradition in computational stylistics shows that linguistic diversity varies systematically across genres, domains, and literary prestige \citep{biber2009register, underwood2016longue}; neural style representations trained to be content-independent now make this variation directly measurable \citep{patel2025styledistance}, which is the representation we adopt for our linguistic-diversity facet.

These lines of work establish that narrative texture---how a story moves through thematic, affective, and stylistic space---is measurable rather than merely impressionistic. 
Recent work has begun applying such metrics to LLM evaluation \citep{sui2026spoileralert, tian2024narratives}, but without tracing how each alignment stage shapes the effect. 
We repurpose these metrics as a diagnostic tool to ask at which stage of post-training narrative compression accumulates.

%% file: Sections/03-Experimental-Design.tex
\section{Experimental Design}
\label{sec:design}

\paragraph{Matched continuation setting.}
We evaluate literary continuation rather than free generation.
For each story, we reveal a prefix at cut point $c \in \{40\%,\;60\%,\;80\%,\;90\%\}$ and treat the remaining original text as the matched human continuation.
Each model receives the same prefix, and all measurements are computed only on the continuation.
The four cut points yield continuations from different narrative positions, testing whether flattening effects are robust across how much context the model has seen.

\paragraph{Story domains.}
We select three corpora that span a range of human creative-writing settings, differing in task framing and pre-publication gatekeeping (Table~\ref{tab:corpora}).

\begin{itemize}[noitemsep, topsep=2pt, leftmargin=*]
    \item \textbf{Professional literary fiction.}
    3,023 short stories published in \textit{The New Yorker} between 
    1945 and 2019 \citep{shaalan2022view}, restricted to stories under 
    5\,K words.
    These stories have passed professional editorial selection and 
    represent a high-density literary baseline with rich thematic 
    modulation, affective tension, and stylistic variety.

    \item \textbf{Prompt-guided common fiction.}
    \textsc{Tell Me A Story} (TMAS) comprises human-written stories
    elicited by explicit creative-writing prompts and produced by a
    fixed pool of 28 recruited writers in guided, multi-week workshops:
    each story passed a draft, peer-feedback, revision, and
    workshop-lead approval gate before inclusion
    \citep[\S4.3]{huot2024agents}.
    Because writers respond to a shared task instruction, this corpus
    captures human fiction produced under conditions that parallel
    instruction-following generation.

    \item \textbf{Public-platform fiction.}
    \textit{StoryStar} stories are drawn from a free online
    self-publishing platform \citep{storystar}.%
    \footnote{\url{https://www.storystar.com}}
    There is no pre-publication review, acceptance criterion, or
    editorial gate; reader ratings apply only after posting.
\end{itemize}

\noindent
The TMAS--\textsc{StoryStar} contrast is one of task framing and process gatekeeping, not writer credential: neither corpus certifies its authors, and both are non-professional relative to \textit{New Yorker} fiction. It is also not load-bearing. The central contrast is professional versus common fiction, and the two common-fiction corpora cluster together on every facet while separating jointly from \textit{The New Yorker} (Table~\ref{tab:domain-baselines}, Figure~\ref{fig:convergence}C); no headline claim depends on the TMAS--\textsc{StoryStar} gap, and the two instead serve as mutual robustness checks within the common-fiction regime. \looseness=-1

\begin{table}[h]
\centering
\small
\setlength{\tabcolsep}{3pt}
\begin{tabularx}{\columnwidth}{l c X}
\toprule
\textbf{Domain} & \textbf{Stories} & \textbf{Characterization} \\
\midrule
\textit{New Yorker}   & 3{,}023 & Professional literary fiction \\
\textit{TMAS}          & 230     & Prompt-guided human fiction \\
\textit{StoryStar}     & 100     & Public-platform fiction \\
\bottomrule
\end{tabularx}
\caption{
Three human story domains used in this study, ordered by pre-publication
gatekeeping. Together they span professional, task-elicited, and open-platform
writing.
}
\vspace{-4mm}
\label{tab:corpora}
\end{table}

\noindent
Together, the three domains let us separate two effects: how post-training reshapes continuation dynamics (Axis~1, training stage), and whether the magnitude of that reshaping depends on the narrative texture of the human source domain (Axis~2, story domain). \looseness=-1

\paragraph{Cross-domain controls.}
To ensure that observed differences reflect narrative properties rather than surface confounds, we apply several controls across corpora.
Stories in all three domains are restricted to a comparable word-length range.
The same four cut points are applied uniformly.
Model continuations use a fixed decoding configuration across stages and domains. Prompting is held constant within model interface: base models receive the raw story prefix, while instruction-tuned models use the tokenizer-provided chat template with the same continuation instruction.

\paragraph{Models and generation.}
We compare four checkpoints from the OLMo 32B instruction path: 
Base, SFT, DPO, and RLVR \citep{olmo2025olmo2}.
These checkpoints share the same base model and form a sequential post-training path, making post-training stage the primary experimental variable.
For each story--cut--model tuple, we sample $5$ continuations under identical decoding settings. The prompt and Model setting are available in Appendix~\ref{app:generation}.
Appendix~\ref{app:prompt-interface-control} reports a prompt-interface control; Base→SFT should be interpreted as interface + post-training, while all others share the instruction interface.

\paragraph{Sentence-level facets.}
We represent each continuation as a sentence-level trajectory.
For a continuation with sentences $x_1,\ldots,x_T$, encoder $f_d$ maps each sentence to a facet-specific vector $z_t^{(d)} = f_d(x_t)$, yielding a trajectory through that facet space. We measure three facets, defined next.

\paragraph{Flattening metrics.}
We operationalize narrative flattening as reduced variation relative to matched human continuations; full metric definitions are in Appendix~\ref{app:metrics} and validity checks for each facet in Appendix~\ref{app:metric-validity}.
For \textbf{thematic motion}, we encode each sentence with \texttt{text-embedding-3-large} (3072d) and measure the coefficient of variation (CV) of sentence-to-sentence semantic jump sizes within each continuation. CV is high when a continuation alternates dwelling (several small jumps) with abrupt shifts (one large jump) and low when it advances in uniformly sized steps: it measures the \emph{evenness} of thematic motion, not its speed.

For \textbf{affective prevalence}, a literary-adapted GoEmotions classifier \citep{demszky-etal-2020-goemotions} (28 classes; Appendix~\ref{app:affect}) assigns each sentence a top-1 emotion label; we track the prevalence of
conflict, surprise--curiosity, and neutral affect.
The three families are chosen on theoretical grounds: structural-affect theory treats surprise, curiosity, and conflict as the engines of narrative tension \citep{sternberg1978expositional, BREWER1982473}. Remaining families are tracked separately and stay comparatively stable, so the compression we report is selective rather than global (Appendix~\ref{app:affect-family-robustness}).
The classifier is validated against human annotations (family-level $\kappa = 0.68$; Appendix~\ref{app:affect}), and its errors under-count high-arousal affect symmetrically for human and model text, so they attenuate rather than manufacture the gaps we report (Appendix~\ref{app:affect-classifier-robustness}).

For \textbf{linguistic diversity}, we represent each story with StyleDistance embeddings (768d) and measure Maximum Mean Discrepancy (MMD) to the human style distribution and across-story variance. These two estimators carry all style claims; the style PCA in Table~\ref{tab:domain-baselines} and Figure~\ref{fig:convergence}C is a separate descriptive projection of the same embeddings, used only to characterize baseline domain differences and never as a test statistic.
Length-sensitive metrics are residualized for continuation length. \looseness=-1

%% file: Sections/04-Results.tex
\section{Results}
\label{sec:results}

\subsection{Human Story Domains Differ in Baseline Narrative Texture}
\label{sec:domain-baselines}

Before examining model continuations, we compare human continuations across the three domains (Table~\ref{tab:domain-baselines}).

\begin{table}[H]
\centering
\small
\setlength{\tabcolsep}{4pt}
\begin{tabular}{lccc}
\toprule
Corpus & Theme CV & Affect \% & Style Axis \\
\midrule
\textit{New Yorker} & 0.105 & 41.0 & 0.22 \\
\textsc{TMAS} & 0.098 & 26.5 & $-$2.07 \\
\textsc{StoryStar} & 0.110 & 23.2 & $-$1.99 \\
\bottomrule
\end{tabular}
\caption{Human story domains differ in thematic rhythm (per-story topic-jump CV), affective charge (high emotional intensity \%), and stylistic register (mean $z$-score on style PCA, which explains 88.5\% of variance; positive = more literary). 95\% bootstrap CIs are reported in Appendix~\ref{app:full-cross-corpus}; all pairwise domain differences in affective charge and style position are significant.}
\label{tab:domain-baselines}
\vspace{-3mm}
\end{table}

In \emph{thematic motion}, the domains differ modestly: topical step-size CV ranges from ${\sim}0.098$ (\textsc{TMAS}) to ${\sim}0.110$ (\textsc{StoryStar}), with \textit{The New Yorker} in between at ${\sim}0.105$.
The contrasts in \emph{affective prevalence} are larger:
\textit{The New Yorker} carries the highest combined surprise--curiosity and conflict share (${\sim}41\%$), with \textsc{TMAS} at ${\sim}26\%$ and \textsc{StoryStar} lowest at ${\sim}23\%$.
\emph{Linguistic diversity} shows the sharpest divide.
On the first principal component of style-neural embeddings (88.5\% of variance; positive = more literary register), \textit{The New Yorker} stories cluster at $z = 0.22$, while \textsc{TMAS} ($-$2.07) and \textsc{StoryStar} ($-$1.99) occupy a distinct, nearly overlapping region more than two standard deviations away.
These results show that the three corpora differ most sharply in affective charge and stylistic signature, with \textit{The New Yorker} separated from the other two corpora, while prompt-guided and public-platform fiction cluster together.
This raises the question that motivates the rest of our analysis: if post-trained continuations fall on the common-fiction side of this divide, the model--human gap should be largest for \textit{The New Yorker}, whose human baseline lies farthest from that side.

\subsection{Post-Training Flattens Story Continuation Across Facets}
\label{sec:posttraining-facets}

\begin{figure*}[t]
    \includegraphics[width=\textwidth]{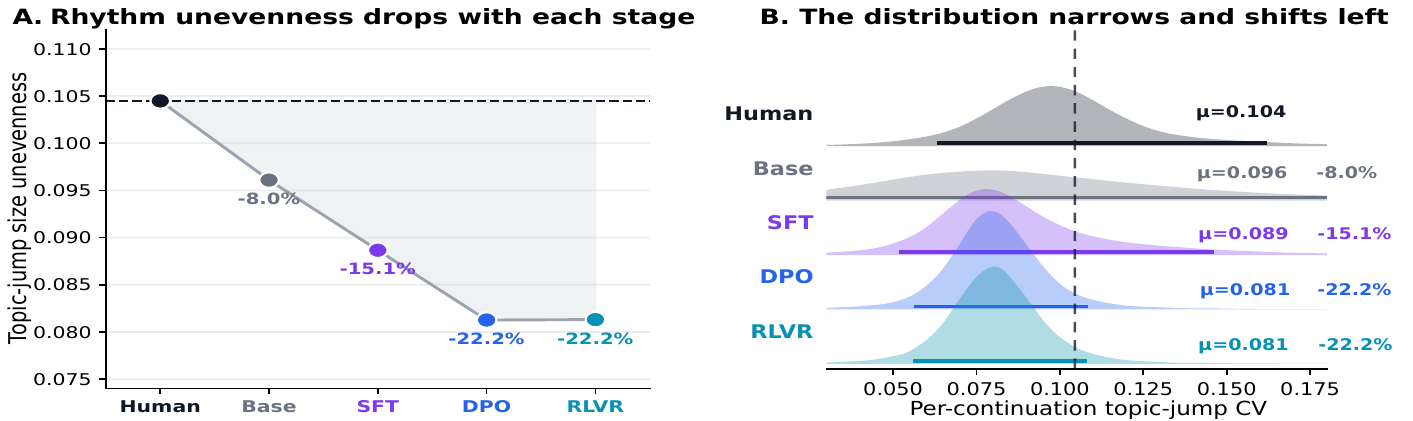}
    \caption{\textbf{(A)} Per-story CV ($\sigma/\mu$) of sentence-to-sentence topic-jump $L_2$ distances. Dashed line = human mean; percentages = unevenness lost relative to human. Length regression confirms continuation length does not confound this metric ($R^{2} < 0.001$).
    \textbf{(B)} Distribution of the same per-continuation CV. Dashed line = human mean; brackets span the 5th--95th percentile. Post-training progressively narrows the distribution and shifts it leftward. 95\% bootstrap CIs are narrower than the plot markers at all stages (Appendix~\ref{app:full-cross-corpus}).}
    \label{fig:stage-progression-topic}
    \vspace{-4mm}
\end{figure*}

\begin{figure*}[t]
    \includegraphics[width=\textwidth]{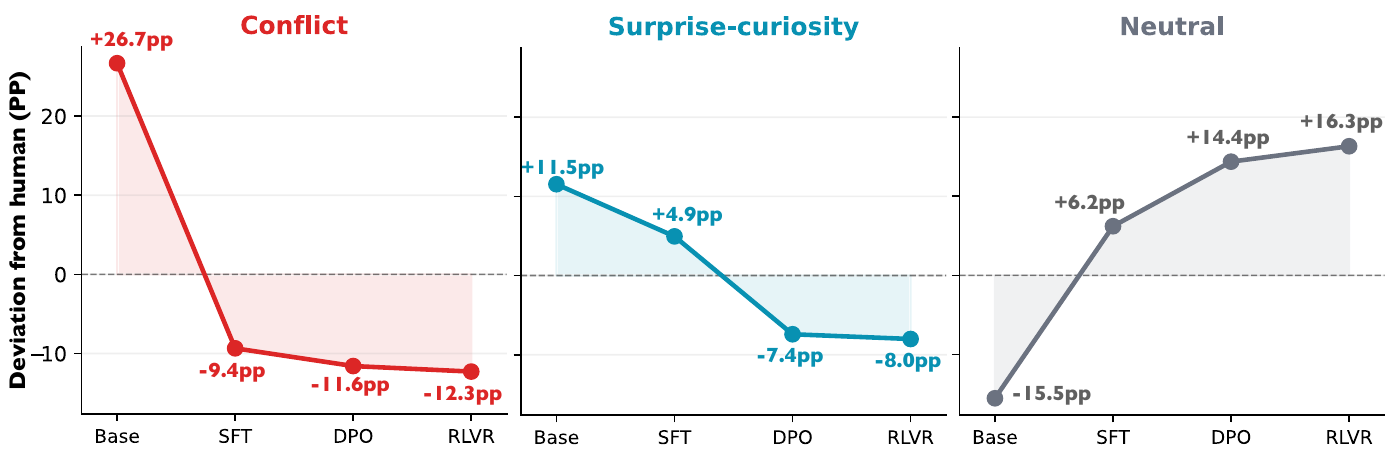}
    \caption{Each panel shows the percentage-point deviation of model prevalence from the matched human baseline; zero on the $y$-axis indicates the human level. Affective prevalence across OLMo post-training stages for
    \textit{The New Yorker} continuations. The base model emerges from pretraining over-marked. Successive post-training stages suppress the focal conflict and surprise--curiosity families and inflate neutral content, overshooting human prevalence rather than converging on it. 95\% story-bootstrap CIs are narrower than the plot markers at all stages (Appendix~\ref{app:full-cross-corpus}); a six-family robustness decomposition is reported in Appendix~\ref{app:affect-family-robustness}.}
    \label{fig:Affective emotion difffernces-NewYorker}
    \vspace{-4mm}
\end{figure*}

Results are shown primarily for \textit{The New Yorker} (Figures~\ref{fig:stage-progression-topic}--\ref{fig:convergence}, Table~\ref{tab:style-progression}); cross-domain comparisons follow in Section~\ref{sec:cross-domain-convergence}, and per-facet breakdowns for the other two corpora are in Appendix~\ref{app:full-cross-corpus}.

\paragraph{Thematic motion.}
The per-story coefficient of variation of topic jumps drops at every stage: from a human mean of ${\sim}0.105$ to $0.096$ at Base ($-8.0\%$), $0.089$ at SFT ($-15.1\%$), and $0.081$ at DPO/RLVR ($-22.2\%$)
(Figure~\ref{fig:stage-progression-topic}A).
The underlying distribution tells the same story: human CV values spread broadly, reflecting stories that mix dwelling with sharp thematic pivots, whereas each post-training stage compresses the distribution leftward into an increasingly narrow peak
(Figure~\ref{fig:stage-progression-topic}B).
A mixed-effects model with story-level random intercepts and fixed cut-point effects confirms the RLVR--human reduction in $L_2$ topical CV ($\beta=-0.0228$, 95\% CI $[-0.0234,-0.0222]$, $p<.001$).
The decline is compressed variation, not slower movement: RLVR's mean adjacent-sentence distance is marginally \emph{larger} than the human mean ($1.167$ vs.\ $1.133$); what falls is the SD of jump sizes ($0.095$ vs.\ $0.117$). \looseness=-1

\paragraph{Affective prevalence.}
The base model emerges from pretraining over-marked: conflict (${\sim}47\%$) and surprise--curiosity (${\sim}33\%$) both exceed human prevalence (${\sim}20\%$ and ${\sim}21\%$ respectively), while neutral content is correspondingly underweighted (${\sim}13\%$ vs.\ ${\sim}29\%$ in humans) (Figure~\ref{fig:Affective emotion difffernces-NewYorker}). 
This likely reflects the over-representation of emotionally marked genres in web-scale pretraining data. 
Post-training does not return the model to human levels; it over-corrects past them. By RLVR, conflict has collapsed to ${\sim}7.5\%$ and surprise--curiosity to ${\sim}13\%$, while neutral content swells to ${\sim}45\%$---each marker now further from the human distribution than the base model was, in the opposite direction. 
Human continuations occupy intermediate, modulated affective values; post-trained outputs selectively compress the two focal high-arousal families measured here while increasing neutral narration. A full family-level decomposition shows that this is not a global collapse of every affect category: sadness/loss and warmth/affiliation remain comparatively stable after SFT (Appendix~\ref{app:affect-family-robustness}).
Flattening effects are directionally stable across all four cut points (Appendix~\ref{app:cutpoint}).
The same model confirms both the reduction in conflict prevalence ($\beta=-0.1228$, 95\% CI $[-0.1270,-0.1185]$, $p<.001$) and the increase in neutral narration ($\beta=0.1632$, 95\% CI $[0.1601,0.1663]$, $p<.001$). These effects remain significant after adding log realized sentence length as a covariate (Appendix~\ref{app:length-covariate}).

\paragraph{Linguistic Diversity.}
MMD to the human style distribution increases monotonically from $0.24$ (Base) to $0.41$ (SFT) to $0.52$--$0.53$ (DPO/RLVR), while across-story style variance drops from ${\sim}6\times$ human at Base to $0.85\times$ at SFT and $0.5$--$0.55\times$ at DPO/RLVR (Table~\ref{tab:style-progression}).
The base model's MMD is low because its stylistic sprawl is broad enough that its aggregate footprint overlaps human writing, but it is writing in too many directions at once. 
Post-training collapses this sprawl onto a narrow attractor that is offset from the human distribution---the model converges on a single voice and applies it across every story.
A fixed-sentence-count control shows that this variance collapse is not an artifact of different realized continuation lengths (Appendix~\ref{app:style-sentence-count-control}).
Holding each prefix fixed, the model's own five samples contract too: within-prefix trace variance falls to $0.83\times$ Base at SFT and $0.37$--$0.43\times$ at DPO/RLVR, so the collapse is not an artifact of setting one model against many writers (Appendix~\ref{app:within-prefix}).

\paragraph{Generalization across model families.}
The same directional pattern holds outside the OLMo lineage. Base-vs-Instruct endpoint checks on Qwen2.5-32B, Llama-3.1-8B, and Gemma-3-12B lower affective charge in all nine family--domain cells (on \textit{The New Yorker}, $67.6\rightarrow38.9$, $68.4\rightarrow35.3$, and $55.1\rightarrow32.8$ points) and reduce thematic CV in every family, though style compression is less uniform (Appendix~\ref{app:additional-models}). These are endpoint comparisons rather than stage-wise replications, so they establish direction but not stage attribution.

\begin{table}[h]
\centering
\small
\setlength{\tabcolsep}{4pt}
\begin{tabular}{ccc}
\toprule
Stage & MMD\textsuperscript{2} ($\uparrow$\,farther) & Var/human ($\downarrow$\,less) \\
\midrule
Human & --- (ref) & 1.00 (ref) \\
Base  & 0.25 & 6.03 \\
SFT   & 0.41 \ua{+64\%} & 0.84 \da{$-$86\%} \\
DPO   & 0.53 \ua{+112\%} & 0.49 \da{$-$92\%} \\
RLVR  & 0.52 \ua{+108\%} & 0.52 \da{$-$91\%} \\
\bottomrule
\end{tabular}
\caption{Style divergence and variance across post-training stages for New Yorker continuations.
Each successive stage pushes the model's style distribution farther from the human reference (higher MMD\textsuperscript{2}) while collapsing across-story variation (lower variance), converging onto a narrow stylistic attractor offset from human writing. MMD confidence intervals use the sentence-state bootstrap described in Appendix~\ref{app:embedding-details}; across-story variance uses story-level bootstrap.}
\label{tab:style-progression}
\vspace{-4mm}
\end{table}

\subsection{Post-Training Erases Cross-Domain Differences, Compressing Professional Fiction Most}
\label{sec:cross-domain-convergence}

\begin{figure*}[htbp]
    \includegraphics[width=\textwidth]{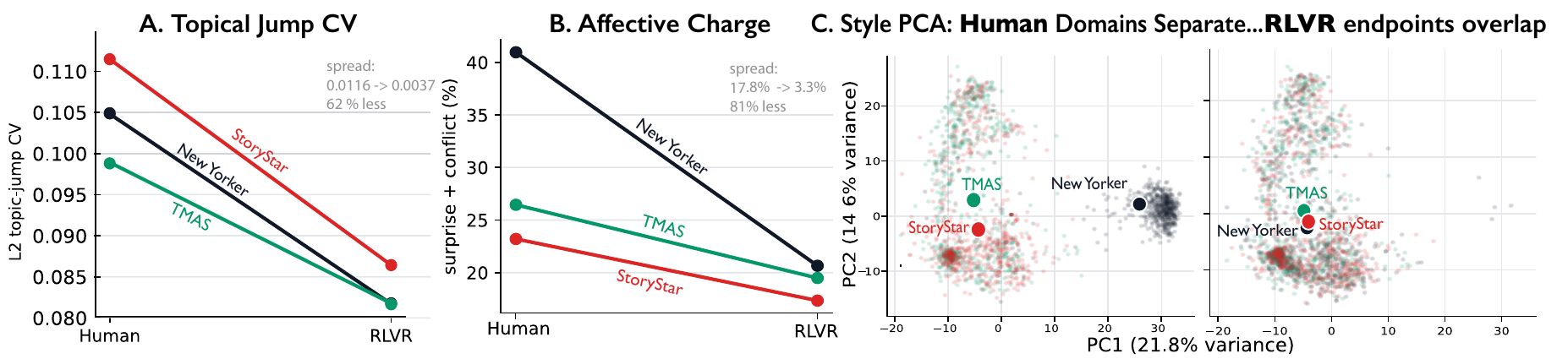}
    \caption{Cross-domain convergence under post-training.
    \textbf{(A)} Domain-mean topical jump CV for three corpora
    (\textit{New Yorker}, \textit{Tell Me A Story}, \textit{StoryStar}) at the
    human and RLVR endpoints; corpus spread shrinks $0.0116 \to 0.0037$
    ($62\%$). 
    \textbf{(B)} Domain-mean affective charge (surprise--curiosity + conflict,
    \%) at the same two endpoints; spread shrinks $17.8 \to 3.3$ percentage
    points ($81\%$).
    \textbf{(C)} PCA of 768-dimensional StyleDistance embeddings. Each small dot is an individual story continuation, color-coded by domain; each large circle is the centroid of one domain (mean position of stories). The left panel shows human-authored continuations; the right shows the same three domains continued by RLVR. Under human authorship, \textit{New Yorker} occupies a clearly separated region of style space while \textsc{TMAS} and \textsc{StoryStar} cluster together; under RLVR, all three centroids collapse onto a single shared region. }
    \label{fig:convergence}
    \vspace{-4mm}
\end{figure*}

Section~\ref{sec:domain-baselines} established that human story domains occupy distinct narrative regimes---\textit{The New Yorker} separated from \textsc{TMAS} and \textsc{StoryStar} by higher affective charge and a markedly different stylistic signature.
We find that after post-training, cross-domain differences in model outputs become substantially smaller than cross-domain differences in human outputs. (Figure~\ref{fig:convergence}). 

\paragraph{Cross-domain spread collapses under post-training.}
Across three measurement facets, the gap between domains narrows though the degree of convergence varies.
In \emph{thematic motion}, the range of topical jump CV drops from $0.0116$ (human) to $0.0037$ (RLVR)---a substantial $62\%$ reduction, but leaving residual cross-domain structure (Figure~\ref{fig:convergence}A).
In \emph{affective prevalence}, surprise+conflict range falls from $17.8$ to $3.3$ percentage points, an $81\%$ reduction (Figure~\ref{fig:convergence}B).
Convergence is strongest in \emph{linguistic habits}: human cross-domain MMD\textsuperscript{2} reaches $0.61$, while the largest RLVR cross-domain MMD\textsuperscript{2} is $0.01$---a substantial reduction (Figure~\ref{fig:convergence}C).
By the final post-training stage, the three RLVR endpoints occupy largely overlapping regions of stylistic space despite originating from sharply different human domains.\looseness=-1

\paragraph{Professional fiction bears the largest gap.}
Because post-training pushes all domains toward a similar continuation regime, the domain that starts farthest from that regime undergoes the largest shift.
In affective prevalence, \textit{The New Yorker}'s combined surprise+conflict share falls from ${\sim}41\%$ to ${\sim}20\%$ at RLVR---a drop of roughly $21$ percentage points, compared with ${\sim}7$ points for \textsc{TMAS} and ${\sim}6$ points for \textsc{StoryStar}.
Linguistic habits provide the starkest evidence (Figure~\ref{fig:convergence}C): under human authorship \textit{The New Yorker} occupies a clearly separated region of style space, while \textsc{TMAS} and \textsc{StoryStar} sit on top of one another; under RLVR, all three collapse onto a shared region that, in the two leading principal components, overlaps with the common-fiction human baselines (full-dimensional MMD in Appendix~\ref{app:mmd_details}).
The smaller gaps for \textsc{StoryStar} and \textsc{TMAS} do not mean the model writes these domains better; they mean these baselines were already closer to where post-training pushes the model. \looseness=-1

\paragraph{A domain-agnostic continuation regime.}
Under this post-training pipeline, model continuations across the three domains converge, with cross-domain differences substantially smaller than those observed in human continuations. As the convergence pattern above shows, the output regime sits closer to common-fiction baselines than to professional literary fiction.
The resulting continuations are coherent, readable, and sequentially well-formed, but they shed the thematic unevenness, affective tension, and stylistic variation that distinguish human storytelling---and that distinguish one kind of human storytelling from another.

%% file: Sections/06-Discussion.tex
\section{Discussion}
\label{sec:discussion}

\paragraph{Post-training as narrative regularization.}
Our results suggest post-training acts as a kind of regularization. This is not the parametric regularization familiar from optimization, but a regularization of the narrative trajectory. The base model is not simply human-like: it can be affectively over-marked and stylistically wide-ranging and unstable. Supervised fine-tuning, preference optimization, and verifiable-reward reinforcement learning make generation more controlled and conventionally well-formed, but this control is achieved by suppressing variation. Thematic jumps become more uniform, affect is muted, and linguistic habits collapse into a narrower attractor. These findings offer one possible structural account of why aligned fiction can feel mechanical: the issue may not be grammar but the loss of dynamic contrast over the course of a continuation. Direct reader studies linking these structural metrics to human perception remain an important next step.

\paragraph{What the cross-domain design reveals.}
Evaluating model fiction against a pooled human reference would blur a distinction the cross-domain design makes visible. Professional literary fiction, prompt-guided fiction, and public-platform fiction differ before any model is introduced: they vary in affective charge, stylistic signature, and narrative rhythm. A single-baseline evaluation against \textsc{StoryStar} alone would suggest the model is largely faithful; against \textit{The New Yorker} alone would suggest catastrophic gaps, and the pooled average masks both. The cross-domain design shows that the human--model gap is not a fixed property of the model but a function of which human baseline is used to measure it, and that post-trained endpoints converge toward each other regardless of the source they were asked to continue.

\paragraph{Geometry, not yet mechanism.}
Our analysis localizes \emph{where} post-trained outputs sit relative to human baselines, but not \emph{why}. Several mechanisms are compatible with the geometry we observe. Preference annotators may reward more conventional continuations; instruction-tuning data may overrepresent task-oriented prose whose style bleeds into open-ended generation; RLVR's verifiability pressure may favor common-fiction-aligned structures; reward models trained on broad preferences may compress the literary tail of the human distribution. Disentangling these channels requires access to the full SFT, DPO, and RLVR mixtures, along with controlled interventions on each, which we leave to future work. The current conclusion is geometric: the post-trained endpoint lies closer to common-fiction baselines than to professional literary fiction, with the gap widening at each training stage.

\paragraph{Flattening or professionalization?}
A natural alternative reading is that what we call flattening is what (muscular) literary editors do. They remove excess affective marking, smooth erratic motion, and enforce stylistic consistency. The base model's affective over-marking (\S\ref{sec:posttraining-facets}) is consistent with this reading: perhaps post-training is editing the model toward a more disciplined voice. Two findings cut against this interpretation. First, the post-trained endpoint does not approach the human distribution; it overshoots in the opposite direction, ending farther from human affect than the base model was, on the muted side. Second, professional human editing produces \textit{The New Yorker}, the domain from which post-trained models move most aggressively away. Post-training is not professionalizing fiction. It is converging on a third regime that is neither the base model's sprawl nor the literary editor's discipline, but a default closer to common-fiction baselines. \looseness=-1

\paragraph{Implications for creative-writing assistants.}
These findings complicate the success criteria for post-training in open-ended generation. In summarization, reduced variation is desirable. Users want consistency. In fiction, variation leads to heightened experience, and uniform compression becomes a bug rather than a feature. Creative-writing assistants may require alignment objectives distinct from general-purpose assistants: domain-conditional reward models that calibrate to the source register rather than a pooled human preference; distributional matching objectives that preserve variation within a continuation as well as across stories, rather than only point-wise quality; or preference data deliberately sampled from the literary tail rather than from majority-preferred continuations. These are directions the diagnostics point to, not solutions we have tested. Our metrics provide a structural account of what \citet{doshi2024generative} observed at the level of collective output. Where they showed that human--AI co-writing reduces diversity across stories, we show which narrative dimensions are compressed and how compression accumulates across training stages. Narrative-flattening metrics complement human-preference evaluation by asking not whether each continuation is plausible, but whether a collection of continuations preserves the variance of human fiction.

%% file: Sections/07-Conclusion.tex
\section{Conclusion}

We compared matched human story continuations with four OLMo 32B checkpoints across three fiction domains. 
Across thematic motion, affective prevalence, and linguistic diversity, post-training regularizes narrative trajectories: topic jumps become more uniform, high-intensity affect gives way to neutral narration, and across-story style variance collapses. 
This compression is strongest for \textit{The New Yorker} because its human baseline lies farthest from the aligned model's continuation regime; \textsc{TMAS} and \textsc{StoryStar} appear closer mainly because their baselines already sit nearer that attractor. 
The result is not generic degradation but domain-insensitive convergence. 
For creative-writing assistants, coherence and readability are therefore insufficient objectives: alignment should preserve source-domain variation, making narrative flattening a measurable target for future training.

%% file: Sections/08-Limitations.tex
\section*{Limitations}

\textbf{The analysis is diagnostic, not mechanistic}. We do not inspect the SFT, DPO, or RLVR mixtures at the granularity needed to attribute compression to particular training signals---reward-model preferences, instruction style, or verifiability pressure. Our conclusions are geometric: they locate where the post-trained endpoint sits relative to human baselines, not why it moves there. Separating these channels would require controlled interventions on the training pipeline that we cannot perform.

\textbf{Reader-level validation is missing}. We measure compression in the distributional structure of model outputs, not whether readers experience flattened continuations as less compelling or more interchangeable. ``Flatness'' is ultimately perceptual, and our metrics operationalize it through structural proxies; we therefore make no claim about story quality. Eliciting reliable judgments at scale for long-form generation is itself difficult \citep{karpinska2021perils}, and closing this gap remains the most important extension of this work.

\textbf{Measurement instruments may carry the bias they measure}. Each facet depends on an encoder trained largely on web prose, which may inflate distances to literary writing. We validate where we can---the affect classifier against human annotation (Appendix~\ref{app:affect}), the thematic metric against an order-shuffled null, and the style collapse against non-neural surface features (Appendix~\ref{app:metric-validity})---but absolute gap magnitudes should still be read with care. The direction of the effect is robust to this concern.

\textbf{Coverage}. Our stage-wise design traces one instruction-tuning path, so OLMo 32B is the appropriate setting for that claim but leaves transfer open; the Base-vs-Instruct checks on three further families (Appendix~\ref{app:additional-models}) are endpoint comparisons and cannot attribute movement to a stage. The three corpora are also all English, short-form, and contemporary, and \textit{The New Yorker} is a single publication whose editorial style is not coextensive with professional literary fiction. Genre fiction, novel-length narrative, screenplay, poetry, and non-English traditions remain untested. \looseness=-1

\textbf{One task, one decoding regime}. The matched continuation paradigm holds prefix, length, and decoding constant, which is what makes the comparison clean, but the prefix also anchors tone and topic; free generation, multi-turn co-writing, and longer horizons may show larger effects, so our estimates should be read as a lower bound. Prior work likewise shows that decoding substantially shapes diversity \citep{holtzman2020curious}, and aggressive sampling may restore some variance without reversing the regime shift.

%% file: Sections/Ethical_Statement.tex
\section*{Ethical Considerations}

\paragraph{Copyright and data release.}
The \textit{New Yorker} corpus is held in a secure university repository and used only for non-commercial, non-distributing research; TMAS and \textsc{StoryStar} are drawn from public-facing sites. We report only aggregate statistics and release no raw stories, long excerpts, or generated continuations conditioned on copyrighted prefixes. Released artifacts are limited to code, aggregate metric tables, and public-domain or synthetic examples (Appendix~\ref{app:ethics}).

\paragraph{Privacy.}
For public-platform stories we remove author names, bylines, source paths, and other identifiers before analysis. The unit of analysis is the story text and its aggregate continuation metrics; we make no claims about individual authors or demographic groups.

\paragraph{Annotation and AI assistance.}
The affect classifier was adapted with LLM-distilled labels and validated on a 1{,}000-sentence subset annotated by one author and two graduate-student volunteers. Annotation was voluntary and unpaid, and we report no annotator-level analysis. We also used LLMs for sentence-level polishing and limited implementation assistance; all LLM-suggested writing and code was reviewed, edited as needed, and verified by the authors.

\paragraph{Potential harms.}
This work is diagnostic rather than prescriptive: we do not claim that professional literary fiction is the only valid style, nor that lower affective charge is inherently worse. The risk we identify is aggregate homogenization---as many writers come to rely on the same aligned systems, diverse storytelling practices may be nudged toward a narrower continuation regime. We offer these metrics as tools for measuring that risk, not for ranking authors or prescribing a standard for good fiction.

%% file: Sections/Appendix.tex
\appendix

\input{Sections/A01-Data-Statement-and-Corpus-Construction}

\input{Sections/A02-Generation-Setup}

\input{Sections/A03-Affective-classifier-adaptation}

\input{Sections/A04-Mapping-GoEmotions-Labels-into-Narrative-Affect-Families}

\input{Sections/A04_1_Affective_Robustness}

\input{Sections/A05-Embedding-Details}

\input{Sections/A06-Metric-Definitions}

\input{Sections/A07-Statistical-Inference}

\input{Sections/A07-2-Metric-Validity}

\input{Sections/A07-1-prompt_interfeces_control}

\input{Sections/A08-Full-Cross-Domain-Result}

\input{Sections/A09-Qwen}

\input{Sections/A10-Cut-point-Robustness}

\input{Sections/A12-Ethics-Statement}

\input{Sections/A11-Qualitative-Illustration}


%% file: Sections/A01-Data-Statement-and-Corpus-Construction.tex
\section{Data Statement and Corpus Construction}
\label{app:data}

\paragraph{Overview.}
We evaluate story continuation across three English short-fiction domains:
professional literary fiction, prompt-guided human fiction, and public-platform
fiction. Table~\ref{tab:app-corpus-stats} reports corpus-level statistics after
filtering.

\begin{table*}[t]
\centering
\small
\setlength{\tabcolsep}{3pt}
\begin{tabular}{lrrrrrr}
\toprule
\textbf{Corpus} & \textbf{Raw stories} & \textbf{Used stories} &
\textbf{Filter} & \textbf{Mean words} & \textbf{Median words} & \textbf{IQR} \\
\midrule
\textit{The New Yorker} & 5001 & 3{,}023 & $<5{,}000$ words & 2{,}763 & 2{,}695 & 1{,}737--3{,}776 \\
\textsc{Tell Me A Story} & 230 & 230 & full dataset & 1{,}468 & 1{,}405 & 1{,}034--1{,}898 \\
\textsc{StoryStar} & 100 & 100 & full collected set & 2{,}614 & 2{,}478 & 1{,}703--3{,}416 \\
\bottomrule
\end{tabular}
\caption{
Corpus statistics after filtering. Word counts are computed after preprocessing
and before prefix truncation. The interquartile range (IQR) reports the 25th--75th
percentile word-count range.
}
\label{tab:app-corpus-stats}
\end{table*}

\paragraph{\textit{The New Yorker}.}
We use a restricted corpus of short fiction published in \textit{The New Yorker}
between 1945 and 2019. To focus on short-form continuation and to maintain a
comparable length range across corpora, we retain stories under 5{,}000 words.
The source texts are stored in a restricted university repository and are used
only for non-commercial research. We do not redistribute the original stories,
long excerpts, or any generated continuations that would reproduce substantial
copyrighted context. All reported results are aggregate statistics.

\paragraph{\textsc{Tell Me A Story}.}
\textsc{Tell Me A Story} (TMAS) is a publicly available corpus of human-written
stories composed in a guided creative-writing setting. We use the full corpus of
230 stories. Because TMAS writers respond to explicit writing prompts, this
domain provides a human baseline for prompt-guided fiction.

\paragraph{\textsc{StoryStar}.}
\textsc{StoryStar} is a corpus of 100 short stories collected from
\texttt{storystar.com}, a public creative-writing platform. Source-level author
identifiers are not used as analytic variables. The corpus is used as a
public-platform fiction baseline with minimal editorial filtering. The scrape
was conducted in February 2026. We do not redistribute raw StoryStar text unless
permitted by the source license or terms.

\paragraph{Preprocessing.}
Across all corpora, we convert each story to a plain-text record, strip leading
and trailing whitespace, and normalize whitespace when constructing prefixes and
continuations. File names, source paths, prompt fields, and collection metadata
are retained as metadata rather than analyzed as narrative text. Titles and bylines are removed before segmentation. 
We do not apply automatic semantic rewriting or
paraphrase-level deduplication; after the file-level corpus filters above, each
remaining file is treated as one story record. Stories are sentence-split using
the same deterministic rule-based splitter used by the continuation-generation
pipeline: sentence boundaries are placed after terminal punctuation
(\texttt{.}, \texttt{!}, or \texttt{?}, optionally followed by closing quotes or
brackets) when followed by whitespace and an uppercase letter, quotation mark,
opening parenthesis, or opening bracket. Prefix cut points are computed over
these sentence sequences, so each prefix ends at a sentence boundary. For a cut
point $c \in \{40\%,60\%,80\%,90\%\}$, the prefix contains the first $c$
proportion of sentences and the matched human continuation is the remaining
sentence suffix.

%% file: Sections/A02-Generation-Setup.tex
\section{Generation Setup}
\label{app:generation}

\paragraph{Acknowledgement.} This work used the DeltaAI advanced computing and data resource at the National Center for Supercomputing Applications through allocation CIS260671 from the Advanced Cyberinfrastructure Coordination Ecosystem: Services \& Support (ACCESS) program \citep{boerner2023access}, which is supported by U.S. National Science Foundation grants \#2138259, \#2138286, \#2138307, \#2137603, and \#2138296.

\paragraph{Task.}
For each story, we generate continuations from prefixes ending at 40\%, 60\%,
80\%, and 90\% of the sentence sequence. The held-out human suffix is treated as
the matched human continuation. For each story--cut--model tuple, we sample five
model continuations.

\paragraph{Models.}
The primary experiments use four checkpoints from the OLMo 32B instruction path:
Base, SFT, DPO, and RLVR. The exact model identifiers are listed in
Table~\ref{tab:app-models}.

\begin{table}[h]
\centering
\small
\begin{tabular}{ll}
\toprule
\textbf{Stage} & \textbf{Model identifier} \\
\midrule
Base & \texttt{allenai/Olmo-3-1125-32B} \\
SFT & \texttt{allenai/Olmo-3.1-32B-Instruct-SFT} \\
DPO & \texttt{allenai/Olmo-3.1-32B-Instruct-DPO} \\
RLVR & \texttt{allenai/Olmo-3.1-32B-Instruct} \\
\bottomrule
\end{tabular}
\caption{Model checkpoints used in the OLMo stage-wise analysis.}
\label{tab:app-models}
\end{table}

\paragraph{Prompting.}
Base models receive the raw story prefix only:
\begin{quote}
\small
\texttt{\{story\_so\_far\}}
\end{quote}

Instruction-tuned models use the tokenizer-provided chat template with the
following system and user messages:
\begin{quote}
\small
\textbf{System:} You are a fiction writer. Continue the story naturally in the
same style and voice. Write only story text -- no commentary, no meta-discussion,
no preamble, no quotation marks around your continuation.

\textbf{User:} Continue this story to its conclusion in approximately
\texttt{\{n\_words\}} words. Maintain the same tone, style, and narrative voice
throughout. Do not summarize or describe what happens -- write the actual story
text as it would appear on the page.

\texttt{STORY SO FAR:}

\texttt{\{story\_so\_far\}}
\end{quote}

Prompting is therefore held constant across domains within each model interface.
Base and instruction-tuned models differ only in the interface required by the
checkpoint: raw completion for base models and chat-template prompting for
instruction-tuned models.

\paragraph{Compute and infrastructure.}
All generation was run on a mix of two NVIDIA H100 and two NVIDIA L40S GPUs using vLLM. We generated continuations for 3{,}500 stories per checkpoint across the OLMo path (four checkpoints: Base, SFT, DPO, RLVR) and the three additional model families (Qwen2.5-32B, Llama-3.1-8B, and Gemma-3-12B, two checkpoints each), at four cut points with five samples per story–cut–model tuple. Throughput depended on model size and hardware: for the 32B models, generation ran at roughly 0.7 stories/min on the paired L40S and 2.4 stories/min on the paired H100; for the 8B model, roughly 2.6 stories/min per L40S and 4.5 stories/min per H100. Sentence-level affect classification, thematic embedding, and StyleDistance encoding were run as separate downstream passes over the archived continuations and are comparatively inexpensive relative to generation.

\paragraph{Decoding.}
All reported OLMo continuations use the same stochastic decoding configuration:
temperature $=1.2$ and nucleus sampling with top-$p=0.95$. The target
continuation length is set to the word count of the held-out human suffix. The
maximum decoding budget is set dynamically as
\[
\texttt{max\_tokens}
=
\left\lfloor
\texttt{target\_words} \times 1.3 \times 1.15
\right\rfloor,
\]
with a minimum of 64 tokens and a maximum of 2048 tokens. We do not specify a
custom stop sequence; generation stops when the model emits EOS or reaches the
token budget. No explicit decoding seed is passed to vLLM. All generated
continuations used in the analysis are archived, and all downstream metrics are
computed deterministically from those archived generations.

\paragraph{Length diagnostics.}
Because prompt-specified length does not guarantee exact realized length, we
report realized continuation length in Table~\ref{tab:app-length}. Length-sensitive
metrics are additionally residualized by realized sentence length
(Appendix~\ref{app:metrics}).

\begin{table*}[t]
\centering
\small
\begin{tabular}{llrrrr}
\toprule
\textbf{Domain} & \textbf{Source} & \textbf{Mean words} & \textbf{Mean sentences} &
\textbf{Human word ratio} & \textbf{Sentence ratio} \\
\midrule
\textit{New Yorker} & Human & 875.1 & 55.6 & 1.00 & 1.00 \\
\textit{New Yorker} & Base & 497.8 & 12.1 & 0.60 & 0.34 \\
\textit{New Yorker} & SFT & 731.5 & 32.3 & 0.94 & 0.81 \\
\textit{New Yorker} & DPO & 669.8 & 43.1 & 0.93 & 1.03 \\
\textit{New Yorker} & RLVR & 642.3 & 43.2 & 0.90 & 1.03 \\
\textsc{TMAS} & Human & 474.4 & 41.1 & 1.00 & 1.00 \\
\textsc{TMAS} & Base & 477.1 & 16.4 & 1.05 & 0.57 \\
\textsc{TMAS} & SFT & 449.6 & 32.4 & 1.00 & 0.90 \\
\textsc{TMAS} & DPO & 460.3 & 34.9 & 1.04 & 0.95 \\
\textsc{TMAS} & RLVR & 447.8 & 34.9 & 1.02 & 0.96 \\
\textsc{StoryStar} & Human & 848.3 & 71.8 & 1.00 & 1.00 \\
\textsc{StoryStar} & Base & 757.1 & 33.6 & 1.00 & 0.69 \\
\textsc{StoryStar} & SFT & 629.5 & 41.4 & 0.85 & 0.76 \\
\textsc{StoryStar} & DPO & 636.7 & 43.1 & 0.93 & 0.84 \\
\textsc{StoryStar} & RLVR & 608.6 & 42.8 & 0.90 & 0.85 \\
\bottomrule
\end{tabular}
\caption{
Realized continuation lengths. Ratios are computed relative to the matched human
suffix length for the same story and cut point.
}
\label{tab:app-length}
\end{table*}

%% file: Sections/A03-Affective-classifier-adaptation.tex
\section{Affective Classifier Adaptation}
\label{app:affect}

\paragraph{Motivation.}
The original GoEmotions labels are derived from Reddit comments, while our target
domain is literary fiction. We therefore adapt a GoEmotions-style classifier to
literary prose before using it to measure affective prevalence in story
continuations.

\paragraph{Adaptation data.}
We construct a 12{,}000-sentence literary affect adaptation set from external
literary prose sources. The set does not include any model-generated
continuations from the main experiment. Labels are produced via LLM distillation: for each sentence, GPT-5.5 assigns top-3 GoEmotions labels, which are converted to graded multi-label
targets. Human annotators reviewed a calibration subset of 1{,}000 sentences
to verify label quality. (i.e., each sentence receives soft supervision over the 28-label inventory). 
The data are split into 10{,}000 training, 1{,}000 validation, and 1{,}000 held-out test sentences.

To assess the reliability of the LLM-distilled affect labels, three annotators—one of the authors and two graduate-student volunteers with relevant humanities/social-science training—independently annotated a 1{,}000-sentence calibration subset using the same family-level label scheme. Annotation was unpaid and conducted on a volunteer basis. Agreement between the human annotations and the GPT-5.5 silver labels was $\kappa = 0.68$
at the family level, supporting the use of the distilled labels for family-level prevalence analysis rather than fine-grained 28-way affect claims.
Annotation instructions and consent. Annotators were asked to assign each sentence to the affect family that best described its dominant expressed affect, using the family definitions in Tables 9--10; ambiguous cases could be marked as other. Before annotation, volunteers were informed that their labels would be used only for aggregate classifier validation and that no annotator-level analysis would be reported.

\paragraph{Annotation format.}
Each sentence receives scores over the original 28-label GoEmotions inventory.
For the main paper, we evaluate sentence-level top-1 affect after mapping the 28
labels into four categories: three focal narrative affect families
(surprise--curiosity, conflict, and neutral) plus a residual \emph{other}
category (Appendix~\ref{app:emotion-mapping}). This four-way scheme directly
mirrors the contrasts analyzed in the main text.

\paragraph{Training details.}
We initialize from \texttt{SamLowe/roberta-base-go\_emotions} and fine-tune on
the literary adaptation set with binary cross-entropy loss
(\texttt{BCEWithLogitsLoss}), learning rate $2\!\times\!10^{-5}$, batch size 32,
for 5 epochs with early stopping on validation loss. Random seed is 20260423.
For the main top-1 prevalence metric used in the paper, we take the
highest-scoring label and map it to the corresponding affect family; no
probability threshold is applied.

\paragraph{Validation.}
We report classifier performance at two granularities. Because our main analyses
operate at the affect-family level, family-level metrics are the primary
validation; full 28-label results are provided for transparency.

\subparagraph{Family-level performance.}
Table~\ref{tab:app-affect-family} reports four-way classification results
(surprise--curiosity, conflict, neutral, other) on the 1{,}000-sentence
held-out test set. At this granularity the classifier achieves macro-F1$\,=0.623$
and accuracy$\,=0.659$. Restricting to the three focal families that drive our
main findings, macro-F1 rises to $0.676$ (weighted F1$\,=0.720$). The neutral
category, which accounts for the largest share of literary prose, is classified
most reliably (F1$\,=0.818$). Surprise--curiosity shows high precision
($0.754$) but moderate recall ($0.531$), meaning the classifier is conservative:
sentences it labels as surprise--curiosity are usually correct, but it misses
some instances. Because this under-counting bias applies symmetrically to human-
and model-generated continuations, comparative prevalence estimates in the main
text are unlikely to be distorted.

\begin{table}[t]
\centering
\small
\setlength{\tabcolsep}{4pt}
\begin{tabular}{lrrrr}
\toprule
\textbf{Family} & \textbf{Prec.} & \textbf{Rec.} & \textbf{F1} & \textbf{Supp.} \\
\midrule
Surprise--curiosity & 0.754 & 0.531 & 0.623 & 179 \\
Conflict & 0.670 & 0.520 & 0.586 & 125 \\
Neutral & 0.851 & 0.788 & 0.818 & 349 \\
Other & 0.677  & 0.646  & 0.661  & 347 \\
\midrule
\multicolumn{5}{l}{\textit{Three focal families}} \\
\quad Macro avg & 0.758 & 0.613 & 0.676 & 653 \\
\quad Weighted avg & 0.790 & 0.666 & 0.720 & 653 \\
\addlinespace
\multicolumn{5}{l}{\textit{Four-way (incl.\ other)}} \\
\quad Macro avg & 0.629 & 0.621 & 0.623 & 1{,}000 \\
\quad Weighted avg & 0.657 & 0.659 & 0.657 & 1{,}000 \\
\quad Accuracy & & & 0.659 & 1{,}000 \\
\bottomrule
\end{tabular}

\caption{
  Family-level classification on the held-out test set. \emph{Three focal
  families}: the affect categories analyzed in the main text.
  \emph{Four-way}: adds the residual \emph{other} category. The classifier is
  conservative on surprise--curiosity (high precision, moderate recall),
  meaning it under-counts rather than over-counts instances.
}
\label{tab:app-affect-family}
\end{table}

\subparagraph{Fine-grained 28-label performance.}
At the original 28-label granularity the classifier obtains mean
AUC$\,=0.910$, top-1 accuracy$\,=0.529$, and top-3 recall$\,=0.780$ on the
same held-out test set. Table~\ref{tab:app-affect-f1} reports per-label
precision, recall, and F1. Several fine-grained GoEmotions labels (e.g.,
\textit{grief}, \textit{pride}, \textit{relief}) are rare in literary prose
and have low support; the family-level aggregation used in our main analyses
absorbs this long-tail variance.

\begin{table*}[t]
\centering
\small
\begin{tabular}{lrrrr}
\toprule
\textbf{Label} & \textbf{Precision} & \textbf{Recall} & \textbf{F1} & \textbf{Support} \\
\midrule
admiration     & 0.235 & 0.333 & 0.276 & 12 \\
amusement      & 0.538 & 0.269 & 0.359 & 26 \\
anger          & 0.444 & 0.250 & 0.320 & 16 \\
annoyance      & 0.324 & 0.436 & 0.372 & 55 \\
approval       & 0.333 & 0.250 & 0.286 & 12 \\
caring         & 0.333 & 0.053 & 0.091 & 19 \\
confusion      & 0.660 & 0.403 & 0.500 & 77 \\
curiosity      & 0.433 & 0.500 & 0.464 & 52 \\
desire         & 0.312 & 0.357 & 0.333 & 14 \\
disappointment & 0.389 & 0.233 & 0.292 & 30 \\
disapproval    & 0.268 & 0.244 & 0.256 & 45 \\
disgust        & 0.333 & 0.222 & 0.267 &  9 \\
embarrassment  & 0.500 & 0.250 & 0.333 &  4 \\
excitement     & 0.429 & 0.429 & 0.429 & 14 \\
fear           & 0.424 & 0.490 & 0.455 & 51 \\
gratitude      & 0.667 & 1.000 & 0.800 &  2 \\
grief          & 0.000 & 0.000 & 0.000 &  4 \\
joy            & 0.750 & 0.500 & 0.600 & 12 \\
love           & 0.500 & 1.000 & 0.667 &  6 \\
nervousness    & 0.167 & 0.125 & 0.143 & 32 \\
optimism       & 0.200 & 0.417 & 0.270 & 12 \\
pride          & 0.500 & 0.111 & 0.182 &  9 \\
realization    & 0.500 & 0.357 & 0.417 & 28 \\
relief         & 0.400 & 0.143 & 0.211 & 14 \\
remorse        & 0.667 & 0.400 & 0.500 & 10 \\
sadness        & 0.484 & 0.719 & 0.579 & 64 \\
surprise       & 0.417 & 0.455 & 0.435 & 22 \\
neutral        & 0.709 & 0.788 & 0.746 & 349 \\
\midrule
Macro average    & 0.426 & 0.383 & 0.378 & 1{,}000 \\
Weighted average & 0.525 & 0.528 & 0.513 & 1{,}000 \\
Top-1 accuracy   &       &       & 0.528 & 1{,}000 \\
Mean AUC         &       &       & 0.910 & 1{,}000 \\
\bottomrule
\end{tabular}
\caption{
  Per-label held-out test results for the literary-adapted affect classifier at
  the original 28-label GoEmotions granularity (GPT-5.5 silver labels). Main
  analyses use the family-level aggregation in
  Table~\ref{tab:app-affect-family}.
}
\label{tab:app-affect-f1}
\end{table*}

%% file: Sections/A04-Mapping-GoEmotions-Labels-into-Narrative-Affect-Families.tex
\section{Mapping GoEmotions Labels into Narrative Affect Families}
\label{app:emotion-mapping}

The classifier outputs the 28-label GoEmotions inventory. For narrative analysis,
we map the fine-grained labels into three focal affect families plus a residual
category:

\begin{table*}[t]
\centering
\small
\begin{tabularx}{\textwidth}{lX}
\toprule
\textbf{Affect family} & \textbf{GoEmotions labels} \\
\midrule
Surprise--curiosity &
\texttt{confusion}, \texttt{curiosity}, \texttt{realization}, \texttt{surprise} \\
Conflict &
\texttt{anger}, \texttt{annoyance}, \texttt{disapproval}, \texttt{disgust} \\
Neutral &
\texttt{neutral} \\
Other (not analyzed) &
\texttt{admiration}, \texttt{amusement}, \texttt{approval}, \texttt{caring},
\texttt{desire}, \texttt{disappointment}, \texttt{embarrassment},
\texttt{excitement}, \texttt{fear}, \texttt{gratitude}, \texttt{grief},
\texttt{joy}, \texttt{love}, \texttt{nervousness}, \texttt{optimism},
\texttt{pride}, \texttt{relief}, \texttt{remorse}, \texttt{sadness} \\
\bottomrule
\end{tabularx}
\caption{
  Mapping from the 28 GoEmotions labels to the four categories used in the main
  analysis. Each fine-grained label maps to exactly one category. The three
  focal families (surprise--curiosity, conflict, neutral) correspond to the
  narrative contrasts analyzed in the paper; remaining labels are grouped into
  \emph{other}.
}
\label{tab:app-emotion-map}
\end{table*}

Let the affect classifier output a 28-dimensional probability vector for
sentence $s$:
\[
p_s = (p_{s,1}, \ldots, p_{s,28}).
\]
For the main affect prevalence analyses, we first assign each sentence its
top-1 GoEmotions label:
\[
\ell_s = \arg\max_j p_{s,j}.
\]
We then map this label into an affect family:
\[
f_s = M(\ell_s),
\]
where $M$ is the mapping in Table~\ref{tab:app-emotion-map}. For a continuation
$c$ with $N_c$ sentences, family prevalence is:
\[
\mathrm{Prev}_k(c)
=
\frac{1}{N_c}
\sum_{s \in c}
\mathbf{1}[f_s = k].
\]
Reported percentages are:
\[
100 \times \mathrm{Prev}_k(c).
\]
The affective-charge metric used in cross-domain analyses is:
\[
\begin{aligned}
\mathrm{AffectiveCharge}(c)
={}&
\mathrm{Prev}_{\mathrm{surprise\text{-}curiosity}}(c)
\\
&+ \mathrm{Prev}_{\mathrm{conflict}}(c).
\end{aligned}
\]

This top-1 formulation ensures that affect-family percentages are interpretable
as sentence shares and sum to one across the four reported categories.

%% file: Sections/A04_1_Affective_Robustness.tex
\section{Affective-Family Robustness}
\label{app:affect-family-robustness}

The main affective analyses focus on surprise--curiosity, conflict, and
neutral narration. In the main text, \emph{conflict} refers to
\texttt{anger}, \texttt{annoyance}, \texttt{disapproval}, and
\texttt{disgust}. To test whether the result depends on this focal grouping,
we recompute prevalence after splitting the residual affect labels into
additional interpretable families. This analysis uses the same sentence-level
classifier outputs as the main text and therefore requires no additional model
generation or classifier inference.

Table~\ref{tab:app-affect-robustness-map} defines the robustness mapping. It
keeps the main-text families, separates threat/anxiety, sadness/loss, and
warmth/affiliation from the residual category, and leaves the remaining labels
in \emph{other}. This avoids forcing every GoEmotions label into a substantive
narrative category while still reporting all top-1 predictions.

\begin{table*}[t]
\centering
\small
\begin{tabularx}{\textwidth}{lX}
\toprule
\textbf{Affect family} & \textbf{GoEmotions labels} \\
\midrule
Surprise--curiosity &
\texttt{confusion}, \texttt{curiosity}, \texttt{realization}, \texttt{surprise} \\
Conflict &
\texttt{anger}, \texttt{annoyance}, \texttt{disapproval}, \texttt{disgust} \\
Threat/anxiety &
\texttt{fear}, \texttt{nervousness} \\
Neutral &
\texttt{neutral} \\
Sadness/loss &
\texttt{sadness}, \texttt{grief}, \texttt{disappointment}, \texttt{remorse} \\
Warmth/affiliation &
\texttt{admiration}, \texttt{approval}, \texttt{caring}, \texttt{gratitude},
\texttt{love}, \texttt{joy} \\
Other/residual &
\texttt{amusement}, \texttt{desire}, \texttt{embarrassment},
\texttt{excitement}, \texttt{optimism}, \texttt{pride}, \texttt{relief} \\
\bottomrule
\end{tabularx}
\caption{
Robustness mapping from the 28 GoEmotions labels to six interpretable affect
families plus residual other. The \emph{Conflict} row matches the main-text
definition; threat/anxiety is separated here to make the broader robustness
decomposition explicit.
}
\label{tab:app-affect-robustness-map}
\end{table*}

Table~\ref{tab:app-affect-full-family-prevalence} reports the full family
prevalence across the OLMo post-training path. The compression is selective,
not a global collapse of every affect category. Surprise--curiosity and
main-text conflict fall sharply by the DPO/RLVR endpoints, while neutral
narration rises. Threat/anxiety, sadness/loss, and warmth/affiliation do not
follow the same pattern: after SFT they remain comparatively stable or increase
slightly. This pattern argues against a cherry-picked affect result. The
classifier tracks multiple families, but the strongest post-training
compression appears in the focal high-arousal families analyzed in the main
text.

\begin{table*}[t]
\centering
\small
\setlength{\tabcolsep}{3pt}
\begin{tabular}{lrrrrr}
\toprule
Affect family & Human & Base & SFT & DPO & RLVR \\
\midrule
Surprise--curiosity & 21.0 [20.6, 21.4] & 32.5 [31.8, 33.2] & 26.0 [25.6, 26.3] & 13.6 [13.4, 13.7] & 13.0 [12.8, 13.1] \\
Conflict & 20.0 [19.6, 20.4] & 46.7 [45.6, 47.8] & 10.6 [10.4, 10.8] & 8.4 [8.2, 8.5] & 7.7 [7.5, 7.8] \\
Threat/anxiety & 8.0 [7.7, 8.2] & 2.1 [2.0, 2.2] & 8.6 [8.4, 8.7] & 10.2 [10.1, 10.4] & 10.3 [10.1, 10.5] \\
Neutral & 28.7 [28.3, 29.2] & 13.2 [12.8, 13.6] & 35.0 [34.7, 35.2] & 43.1 [42.8, 43.4] & 45.1 [44.7, 45.4] \\
Sadness/loss & 8.6 [8.3, 8.8] & 2.0 [1.9, 2.0] & 9.0 [8.9, 9.2] & 11.0 [10.8, 11.2] & 10.9 [10.7, 11.1] \\
Warmth/affiliation & 6.4 [6.2, 6.6] & 1.5 [1.4, 1.6] & 4.5 [4.4, 4.6] & 5.7 [5.6, 5.8] & 5.6 [5.5, 5.7] \\
Other/residual & 7.3 [7.1, 7.5] & 2.0 [2.0, 2.1] & 6.4 [6.3, 6.5] & 8.0 [7.9, 8.1] & 7.5 [7.4, 7.6] \\
\bottomrule
\end{tabular}
\caption{
Full affect-family prevalence across OLMo post-training stages for
\textit{The New Yorker} continuations. Values are sentence-share percentages,
reported as mean [95\% story-bootstrap CI]. Families are computed by taking the
top-1 GoEmotions label for each sentence and then mapping labels into the
families in Table~\ref{tab:app-affect-robustness-map}. The \emph{Conflict}
row matches the main-text conflict definition; \emph{Threat/anxiety} is
separated here to avoid changing the meaning of the main-text affective-charge
metric.
}
\label{tab:app-affect-full-family-prevalence}
\end{table*}

We also test broader affective-charge definitions. The first column of
Table~\ref{tab:app-affect-expanded-charge} exactly matches the main-text
definition: surprise--curiosity plus conflict. The second column additionally
includes threat/anxiety. The third column further adds sadness/loss and
high-arousal positive labels (\texttt{excitement}, \texttt{joy},
\texttt{amusement}, and \texttt{desire}). This check asks whether affective
charge merely moves from the focal families into other emotionally marked
labels after post-training.

\begin{table*}[t]
\centering
\small
\setlength{\tabcolsep}{3pt}
\begin{tabular}{lcccc}
\toprule
Stage & Main-text charge & Threat-inclusive & Expanded charge & Neutral \\
\midrule
Human & 41.0 [40.5, 41.4] & 48.9 [48.4, 49.4] & 62.5 [62.0, 62.9] & 28.7 [28.3, 29.2] \\
Base & 79.2 [78.6, 79.7] & 81.3 [80.8, 81.8] & 84.8 [84.4, 85.2] & 13.2 [12.9, 13.6] \\
SFT & 36.6 [36.2, 36.9] & 45.1 [44.8, 45.5] & 58.2 [57.9, 58.5] & 34.9 [34.7, 35.2] \\
DPO & 22.0 [21.7, 22.2] & 32.2 [32.0, 32.5] & 47.6 [47.3, 47.9] & 43.1 [42.8, 43.4] \\
RLVR & 20.7 [20.4, 20.9] & 30.9 [30.6, 31.2] & 45.9 [45.7, 46.3] & 45.1 [44.7, 45.4] \\
\addlinespace
RLVR--Human & $-20.3$ & $-18.0$ & $-16.5$ & $+16.3$ \\
\bottomrule
\end{tabular}
\caption{
Alternative affective-charge aggregations. \emph{Main-text charge} is
surprise--curiosity plus conflict, where conflict is \texttt{anger},
\texttt{annoyance}, \texttt{disapproval}, and \texttt{disgust}.
\emph{Threat-inclusive} additionally includes \texttt{fear} and
\texttt{nervousness}. \emph{Expanded charge} further includes sadness/loss
and high-arousal positive labels. All aggregations yield the same qualitative
conclusion: affective marking falls by the RLVR endpoint while neutral
narration rises. Values are mean [95\% story-bootstrap CI]; the final row
reports the RLVR--Human difference in percentage points.
}
\label{tab:app-affect-expanded-charge}
\end{table*}

The expanded aggregation supports the same conclusion as the main affective
analysis. Post-training does not merely move affect from the focal families into
other high-intensity categories; it reduces broad affective marking while
increasing neutral narration.

\subsection{Is the Affective Gap a Classifier Artifact?}
\label{app:affect-classifier-robustness}

The affective facet rests on a classifier whose family-level macro-F1 is
$0.676$ (Appendix~\ref{app:affect}). Two checks ask whether the Human--RLVR gap
could be produced by that error rather than measured through it.

\paragraph{Soft-probability rescoring.}
The main analysis assigns each sentence its top-1 label, so a gap could in
principle reflect where the decision boundary falls rather than the underlying
distribution. We therefore recompute prevalence from the archived 28-dimensional
probability vectors, normalized per sentence and summed by family, discarding
the hard labels entirely. On \textit{The New Yorker} the headline gap survives:
human affective charge is $41.1\%$ [40.8, 41.5] against RLVR at $24.2\%$
[24.0, 24.3], compared with $41.0\%$ and $20.7\%$ under top-1 labelling. The gap
narrows slightly but is not a top-1 artifact.

\paragraph{Classifier-noise simulation.}
We next ask what the measured gap would look like if a given true gap were
observed \emph{through} a classifier with our error profile. Passing the
observed human and RLVR prevalence vectors through the held-out four-family
confusion matrix, at the actual sentence counts, turns a true gap of roughly
$20$\,pp into a simulated observed gap of $10.3$\,pp [10.2, 10.4]: noise of the
measured magnitude attenuates the contrast by about half. Running the
correction in the other direction, constrained matrix inversion recovers true
gaps at least as large as those observed in all three domains
(\textit{New Yorker} $20.3 \rightarrow 41.3$\,pp; \textsc{TMAS}
$6.7 \rightarrow 11.3$; \textsc{StoryStar} $5.6 \rightarrow 8.9$).

Because the same classifier is applied symmetrically to human and model text,
its error can only push the two sources toward each other. Classifier noise can
therefore cause us to \emph{under}-report narrative flattening; it cannot
manufacture it. The affective gaps in the main text should be read as
conservative.

%% file: Sections/A05-Embedding-Details.tex
\section{Embedding Details}
\label{app:embedding-details}

\subsection{Thematic Embeddings}

We encode each sentence using \texttt{openai/text-embedding-3-large} through an
OpenAI-compatible OpenRouter endpoint. Embeddings were generated in
April--May 2026 using model route \texttt{openai/text-embedding-3-large}. No
date-versioned provider snapshot was pinned. We request 3072-dimensional
embeddings directly from the API using \texttt{dimensions=3072}; this is not a
post-hoc PCA or dimensionality reduction step. Returned embeddings are
L2-normalized row-wise before storage and are treated as unit-sphere vectors.
The stored columns are \texttt{top\_0} through \texttt{top\_3071}.

For thematic motion, we encode each sentence with
\texttt{text-embedding-3-large}, requesting 3072-dimensional embeddings
from the API (\texttt{dimensions=3072}). Returned embeddings are
$L_2$-normalized before storage. Sentence-to-sentence movement is
computed using Euclidean distance:
\[
d_{L_2}(z_t, z_{t-1}) = \|z_t - z_{t-1}\|_2.
\]
The main analyses use $L_2$ distance between normalized embeddings; cosine
distance is reported only as a robustness variant.

\subsection{StyleDistance Embeddings}

We encode each sentence with \texttt{StyleDistance/styledistance}, using a local
HuggingFace snapshot
\texttt{b7df5f0b0480773c097ba3121d83ca32b71015ca}. The model is a
SentenceTransformer wrapper over \texttt{FacebookAI/roberta-base} with hidden
size 768. Sentence embeddings are produced by mean pooling over token embeddings,
not CLS pooling. Embeddings are L2-normalized at inference using
\texttt{normalize\_embeddings=True}. The resulting 768-dimensional vectors are
stored as \texttt{styleN\_0} through \texttt{styleN\_767}.

Style MMD is computed on sentence-level 768-dimensional embeddings. Across-story
style variance and PCA homogenization analyses first aggregate sentence
embeddings into continuation centroids by taking the arithmetic mean within each
\[
(\texttt{story\_id}, \texttt{source}, \texttt{position}, \texttt{sample\_id})
\]
group. We do not re-normalize centroids after averaging.

\paragraph{Style MMD.}
\label{app:mmd_details}
We compute MMD using an unbiased MMD$^2$ estimator with a Gaussian/RBF kernel:
\[
k(x,y)=\exp\left(-\frac{d_{\cos}(x,y)^2}{2\sigma^2}\right),
\]
where $d_{\cos}$ is cosine distance in the style-neural embedding space. The
bandwidth $\sigma$ is chosen separately for each comparison using the median
heuristic over pairwise distances in the combined sampled human/model vectors.
It is not cross-validated and is not fixed globally. The main MMD function
subsamples up to 1{,}000 sentence vectors per group before computing MMD. The
cross-corpus style plots first draw up to 2{,}500 candidate sentence vectors per
group, but the MMD computation itself again subsamples to 1{,}000 vectors per
pair. For the main human--model MMD curves, confidence intervals are computed
with a sentence-vector bootstrap after this subsampling step. Because sentence
vectors from the same story are not independent, these intervals should be
interpreted as uncertainty for the sampled sentence-state distribution rather
than as story-block confidence intervals. The cross-corpus style heatmaps are
reported as point estimates without bootstrap confidence intervals.

\paragraph{Style PCA.}
For the two-dimensional style PCA shown in the main text, we sample 600
sentence-level style vectors per stage and domain. PCA is fit on all sampled
human and RLVR points together across the three domains. Before PCA, dimensions
are standardized using a StandardScaler fit on the full sampled matrix. PCA
centroids are simple means of projected points within each stage--domain group.

For the full-population PCA50 homogenization analysis, we first compute
continuation centroids from sentence-level style embeddings. We then z-score each
dimension using the human centroid mean and standard deviation and fit PCA with
50 components on the resulting centroid matrix. Across-story variance is computed
on continuation centroids and bootstrapped by story ID.

\subsection{Style Collapse in Interpretable Surface Features}
\label{app:style-surface-features}

StyleDistance is trained on human-authored text, so a reader may reasonably ask
whether its representation transfers to model-generated prose, and whether the
variance collapse we report is a property of the learned space rather than of
the writing. We therefore repeat the across-story variance analysis on
non-neural surface features that require no learned encoder: comma rate,
function-word ratio, mean word length, type--token ratio, and dialogue-marker
rate, each computed directly from the continuation text and normalized by the
matched human value.

The collapse reproduces. Across-story variance at RLVR is $0.73\times$ the human
level in these interpretable features, and 6 of the 7 features measured
homogenize relative to human continuations. The direction and approximate
magnitude therefore do not depend on StyleDistance: stylistic flattening is
visible in plain surface statistics as well as in learned representations. The
neural estimator remains our primary instrument because it is content-independent
by construction, but it is not carrying the result on its own.

%% file: Sections/A06-Metric-Definitions.tex
\section{Metric Definitions}
\label{app:metrics}

Let a continuation be a sequence of sentences
$Y=(x_1,\ldots,x_T)$. For each facet $d$, sentence encoder $f_d$ produces
\[
z_t^{(d)} = f_d(x_t).
\]

\paragraph{Thematic motion.}
For thematic embeddings, we compute sentence-to-sentence jump sizes:
\[
\Delta_t^{\mathrm{theme}}
=
d(z_t^{\mathrm{theme}}, z_{t-1}^{\mathrm{theme}}),
\qquad t=2,\ldots,T,
\]
where d is $L_2$ distance for the main analyses. Cosine-distance variants are
reported as robustness checks.
\[
\mathrm{CV}_{\mathrm{theme}}
=
\frac{
\mathrm{sd}(\Delta_2^{\mathrm{theme}},\ldots,\Delta_T^{\mathrm{theme}})
}{
\mathrm{mean}(\Delta_2^{\mathrm{theme}},\ldots,\Delta_T^{\mathrm{theme}})
}.
\]
Higher values indicate alternation between dwelling and sharp thematic shifts;
lower values indicate more uniformly sized movement.

\paragraph{Affective prevalence.}
Affective prevalence is computed using the top-1 affect-family formulation in
Appendix~\ref{app:emotion-mapping}. For family $k$ and continuation $c$:
\[
\mathrm{Prev}_k(c)
=
\frac{1}{N_c}
\sum_{s \in c}
\mathbf{1}[f_s = k].
\]
We define $\mathrm{AffectiveCharge}(c)$ as the sum of
$\mathrm{Prev}_{\mathrm{surprise\text{-}curiosity}}(c)$ and
$\mathrm{Prev}_{\mathrm{conflict}}(c)$.

\paragraph{Linguistic-habit distance.}
Let $H=\{h_i\}_{i=1}^{n}$ be sentence-level style embeddings from human
continuations and $M=\{m_j\}_{j=1}^{k}$ the corresponding model embeddings. We
compute unbiased MMD$^2$:
\[
\begin{aligned}
\mathrm{MMD}^2(H,M)
={}&
\frac{1}{n(n-1)}\sum_{i\neq i'} k(h_i,h_{i'})
\\
&+
\frac{1}{k(k-1)}\sum_{j\neq j'} k(m_j,m_{j'})
\\
&-
\frac{2}{nk}\sum_{i,j} k(h_i,m_j).
\end{aligned}
\]
Lower MMD means the model and human style distributions are closer.

\paragraph{Across-story linguistic variance.}
For each continuation, sentence-level style embeddings are averaged into a
continuation centroid $\bar{z}_{s,c,r,m}^{\mathrm{ling}}$. For model stage $m$,
across-story variance is:
\[
V_m
=
\mathrm{tr}
\left(
\mathrm{Cov}
\left(
\left\{
\bar{z}_{s,c,r,m}^{\mathrm{ling}}
\right\}
\right)
\right).
\]
We report this value relative to matched human variance:
\[
V_m^{\mathrm{rel}}=\frac{V_m}{V_{\mathrm{human}}}.
\]
Values below 1 indicate less story-to-story variation than human continuations.

\paragraph{Length residualization.}
For length-sensitive metrics, we fit:
\[
q_i = \alpha + \beta T_i + \epsilon_i,
\]
where $q_i$ is the raw metric for continuation $i$ and $T_i$ is realized
sentence length. We compare stages using:
\[
q_i^{\mathrm{res}} = \hat{\epsilon}_i + \bar{q}.
\]
This removes linear length effects while preserving the original metric scale.

%% file: Sections/A07-Statistical-Inference.tex
\section{Statistical Inference}
\label{app:statistics}

\paragraph{Bootstrap confidence intervals.}
Unless otherwise specified, confidence intervals are computed by nonparametric
bootstrap resampling at the story level. Story-level resampling preserves the
dependence among cut points and generated samples from the same story. For each
bootstrap replicate, we resample stories with replacement, recompute the target
metric, and report percentile 95\% confidence intervals.

\paragraph{Mixed-effects models.}
We fit linear mixed-effects models to verify that the main compression effects
are not artifacts of repeated measurements from the same story. For Human-vs-model
stage-wise checks, we use:
\[
q_{i}
=
\alpha
+
\beta_{\mathrm{stage}(i)}
+
\gamma_{\mathrm{cut}(i)}
+
u_{\mathrm{story}(i)}
+
\epsilon_i,
\]
where $q_i$ is the continuation-level metric, stage is a categorical fixed
effect, cut point is a fixed effect, and $u_{\mathrm{story}(i)}$ is a story-level
random intercept.

For generated-only stage-trend checks, we remove human continuations and model
stage as an ordered variable:
\[
\begin{aligned}
q_i
={}&
\alpha
+ \beta\,\mathrm{stageOrder}_i
+ \gamma_{\mathrm{cut}(i)}
\\
&+ \eta_{\mathrm{sample}(i)}
+ u_{\mathrm{story}(i)}
+ \epsilon_i.
\end{aligned}
\]
Here \(\mathrm{stageOrder}\) indexes Base, SFT, DPO, and RLVR in post-training
order, and sample ID is included as a fixed nuisance effect. We do not include
sample ID in Human-vs-model contrasts because the human continuation has
\(\texttt{sample\_id}=0\), which is structurally confounded with source.

For domain-compression checks, we fit:
\[
\begin{aligned}
q_i
={}&
\alpha
+ \beta_{\mathrm{stage}(i)}
+ \delta_{\mathrm{domain}(i)}
\\
&+ \theta_{\mathrm{stage}(i)\times \mathrm{domain}(i)}
+ \gamma_{\mathrm{cut}(i)}
\\
&+ u_{\mathrm{story}(i)}
+ \epsilon_i.
\end{aligned}
\]
This model tests whether Human-to-RLVR compression differs by source domain.

\paragraph{Multiple comparisons.}
For confirmatory tests within each metric family, we apply Holm-Bonferroni
correction. The headline effects reported in the main text remain significant
after correction. Exploratory robustness tests are reported separately and are
not used as primary evidence.

\begin{table*}[t]
\centering
\small
\begin{tabular}{llrrr}
\toprule
\textbf{Metric} & \textbf{Contrast} & \textbf{Estimate} & \textbf{95\% CI} & \textbf{$p$} \\
\midrule
Topical CV, cosine & RLVR -- Human & $-0.03573$ & $[-0.03672, -0.03474]$ & $<.001$ \\
Topical CV, $L_2$ & RLVR -- Human & $-0.02280$ & $[-0.02338, -0.02222]$ & $<.001$ \\
Topical CV, generated-only trend & stage order & $-0.00848$ & $[-0.00870, -0.00827]$ & $<.001$ \\
Conflict prevalence & RLVR -- Human & $-0.12277$ & $[-0.12703, -0.11852]$ & $<.001$ \\
Neutral prevalence & RLVR -- Human & $0.16320$ & $[0.16012, 0.16628]$ & $<.001$ \\
Style trajectory CV proxy & RLVR -- Human & $-0.13604$ & $[-0.14192, -0.13015]$ & $<.001$ \\
\bottomrule
\end{tabular}
\caption{
Mixed-effects robustness checks. The style row uses a scalar style-trajectory
rhythm proxy; the primary style-distance result remains the sentence-level
StyleDistance MMD and continuation-centroid variance analysis.
}
\label{tab:app-mixed-effects}
\end{table*}

\begin{table}[h]
\centering
\footnotesize
\setlength{\tabcolsep}{3pt}
\begin{tabular}{lrrr}
\toprule
\textbf{Metric} & \textbf{New Yorker} & \textbf{TMAS} & \textbf{StoryStar} \\
\midrule
Topical CV & 0.0356 & 0.0264 & 0.0386 \\
$L_2$ topical CV & 0.0228 & 0.0166 & 0.0247 \\
Affect.\ charge (pp) & 20.31 & 6.65 & 5.61 \\
\bottomrule
\end{tabular}
\caption{
Domain-level Human-to-RLVR compression. Positive values mean Human $>$ RLVR.
New Yorker shows the largest affective-charge compression. For topical rhythm,
New Yorker and StoryStar both show larger compression than TMAS; the
New-Yorker-vs-StoryStar difference is not statistically decisive.
}
\label{tab:app-domain-compression}
\end{table}
\subsection{Length-Covariate Robustness}
\label{app:length-covariate}

Prompt-specified continuation length does not guarantee identical realized
length across stages (Appendix~\ref{app:generation}). To verify that the main
effects are not artifacts of realized continuation length, we refit the main
thematic and affective mixed-effects models with log realized sentence length as
an additional covariate:
\[
\begin{aligned}
q_i
={}&
\alpha
+ \beta_{\mathrm{stage}(i)}
+ \gamma_{\mathrm{cut}(i)}
\\
&+ \lambda \log(1+n_i)
+ u_{\mathrm{story}(i)}
+ \epsilon_i,
\end{aligned}
\]
where \(n_i\) is the number of realized continuation sentences. Stage and cut
point are fixed effects, and story ID is a random intercept.

\begin{table*}[t]
\centering
\small
\begin{tabular}{lrrr}
\toprule
\textbf{Metric} & \textbf{RLVR -- Human} & \textbf{95\% CI} & \textbf{$p$} \\
\midrule
Topical CV, $L_2$ & $-0.02340$ & $[-0.02398, -0.02282]$ & $<.001$ \\
Surprise--curiosity & $-0.06889$ & $[-0.07255, -0.06522]$ & $<.001$ \\
Conflict prevalence & $-0.14425$ & $[-0.14783, -0.14067]$ & $<.001$ \\
Neutral prevalence & $0.17086$ & $[0.16789, 0.17383]$ & $<.001$ \\
\bottomrule
\end{tabular}
\caption{
Length-covariate robustness checks. Models include story-level random
intercepts, cut-point fixed effects, stage fixed effects, and log realized
continuation sentence length. Estimates are on the original metric scale:
topical CV for the thematic row and proportions for affective rows.
}
\label{tab:app-length-covariate}
\end{table*}

The headline contrasts retain the same direction and remain significant after
controlling for realized continuation length.

\subsection{Style Sentence-Count Control}
\label{app:style-sentence-count-control}

The style-variance analysis represents each continuation by the centroid of its
sentence-level style-neural states. Because realized continuation length differs
across stages, a reviewer might worry that variance collapse is partly an
artifact of estimating centroids from different numbers of sentences. We
therefore recompute the variance analysis under a fixed-sentence-count control.

For each continuation, we estimate the expected centroid obtained from a fixed
sample of $K=8$ sentence-level style-neural embeddings. We then recompute
across-story/cut style variance and normalize by the corresponding human
fixed-$K$ variance. Confidence intervals are story-bootstrap intervals. This
control uses the same archived sentence-level style embeddings as the main
analysis and does not require new generation or model inference.

\begin{table}[t]
\centering
\small
\begin{tabular}{lrr}
\toprule
Stage & Sentences/cont. & Fixed-$K$ Var/Human \\
\midrule
Human & 54.7 & 1.00 [0.93, 1.07] \\
Base & 6.2 & 2.50 [2.48, 2.52] \\
SFT & 30.7 & 0.44 [0.43, 0.45] \\
DPO & 42.0 & 0.27 [0.27, 0.28] \\
RLVR & 41.9 & 0.29 [0.29, 0.30] \\
\bottomrule
\end{tabular}
\caption{Style sentence-count control. For each continuation, the style
centroid is recomputed as the expected centroid from a fixed sample of $K=8$
sentence-level style-neural embeddings. Values report across-story/cut style
variance normalized by the corresponding human fixed-$K$ variance.}
\label{tab:app-style-sentence-count-control}
\end{table}

The control changes the magnitude of the base-model estimate, which is expected
because base continuations are much shorter. However, it does not explain the
post-training collapse: SFT, DPO, and RLVR remain far below the human style
variance even when every continuation is placed on the same effective
sentence-count footing. Thus the style-variance result is not an artifact of
using more sentences to estimate human or RLVR centroids.

%% file: Sections/A07-2-Metric-Validity.tex
\section{Metric-Validity Checks}
\label{app:metric-validity}

The thematic facet is a coefficient of variation over adjacent-sentence
semantic distances. Two readings could deflate it without any loss of narrative
structure: the metric might reflect properties of the embedding geometry rather
than sentence \emph{order}, or a lower CV might simply mean the continuation
takes uniformly smaller semantic steps, i.e.\ that it is more locally coherent.
The checks below separate flattening from both.

\subsection{Within-Continuation Shuffle Null}
\label{app:shuffle-null}

For each continuation we hold its sentence embeddings fixed and permute
sentence order 200 times, recomputing CV on every permutation. The permutation
preserves the sentence inventory and the marginal embedding geometry exactly and
destroys only narrative order, so the gap between the observed CV and this null
isolates ordered rhythmic structure. If a low CV reflected encoder geometry, the
observed and shuffled values would coincide.

\begin{table}[t]
\centering
\small
\setlength{\tabcolsep}{4pt}
\begin{tabular}{lrrr}
\toprule
Source & Observed CV & Shuffled null & $\Delta$ \\
\midrule
Human & 0.104 & 0.081 & 0.023 \\
RLVR  & 0.081 & 0.076 & 0.005 \\
\bottomrule
\end{tabular}
\caption{Within-continuation shuffle null for \textit{New Yorker}
continuations. $\Delta$ is the ordered-rhythm excess of the observed CV over a
null that permutes sentence order within the same continuation. Human
continuations carry rhythmic structure beyond their sentence inventory; RLVR
retains little of it. Confidence intervals are given in the text.}
\label{tab:app-shuffle-null}
\end{table}

Table~\ref{tab:app-shuffle-null} reports the result. Human continuations exceed
their own null by $\Delta = 0.023$ [0.022, 0.024], whereas RLVR exceeds its null
by only $0.005$. Fitting the main-text mixed-effects specification to this
excess ($\Delta \sim \text{stage} + \text{cut point} + (1\mid\text{story})$)
gives a Human-vs-RLVR contrast of $\beta = -0.0179$
(95\% CI $[-0.0183, -0.0176]$, $p<.001$), with the same direction on
\textsc{TMAS} ($\beta = -0.0112$) and \textsc{StoryStar} ($\beta = -0.0098$),
both $p<.001$. Because the null is constructed from each continuation's own
embeddings, encoder geometry is identical on both sides of the comparison and
cannot account for the difference.

\subsection{Numerator/Denominator Decomposition}
\label{app:cv-decomposition}

A lower CV could in principle arise from uniformly smaller semantic steps rather
than from compressed variation in step size. It does not.

\begin{table}[t]
\centering
\small
\setlength{\tabcolsep}{4pt}
\begin{tabular}{lrr}
\toprule
Quantity & Human & RLVR \\
\midrule
Mean jump size $\mu$ & 1.133 & 1.167 \\
SD of jump size $\sigma$ & 0.117 & 0.095 \\
\bottomrule
\end{tabular}
\caption{Decomposition of the thematic CV into its numerator and denominator for
\textit{New Yorker} continuations. RLVR takes slightly \emph{larger} average
semantic steps than human text; what falls is the spread of step sizes.
Confidence intervals are given in the text.}
\label{tab:app-cv-decomposition}
\end{table}

RLVR's mean adjacent-sentence distance is marginally \emph{larger} than the
human mean (1.167 vs.\ 1.133); the quantity that falls is the standard deviation
of jump sizes (0.095 vs.\ 0.117). The CV decline is therefore compression of
variation in step size, not a uniform shortening of steps, and it is not
explained by the model advancing more cautiously. Distance function and
continuation length are controlled separately
(Table~\ref{tab:app-mixed-effects}; Appendix~\ref{app:length-covariate}).

Taken together with the shuffle null, this reading also inverts the
local-coherence objection. Aligned continuations are more locally coherent, and
that smoothness \emph{is} the low CV; the higher human CV comes from deliberate
discontinuity---unbridged time-skips, interruptions, register breaks
(Appendix~\ref{app:qualitative})---rather than from incoherence.

\subsection{Within-Prefix Sample Diversity}
\label{app:within-prefix}

The across-story style-variance comparison sets one model against a
multi-author human corpus. To remove the writer-count asymmetry entirely, we
also measure diversity \emph{within} the model's own output: holding each prefix
fixed, we compare the five sampled continuations that stage generates for it,
using identical decoding settings at every stage and the continuation-level
style centroids of Table~\ref{tab:style-progression}.

\begin{table}[t]
\centering
\small
\setlength{\tabcolsep}{4pt}
\begin{tabular}{lrr}
\toprule
Stage & Sample trace var. & Mean pairwise dist. \\
\midrule
Base & 1.00 (ref) & 1.00 (ref) \\
SFT  & 0.83 & 1.14 \\
DPO  & 0.37 & 0.73 \\
RLVR & 0.43 & 0.76 \\
\bottomrule
\end{tabular}
\caption{Within-prefix diversity among the model's own five samples, relative to
Base. Prefix, model lineage, and decoding are held fixed, so no comparison to
human writers is involved.}
\label{tab:app-within-prefix}
\end{table}

Relative to Base, within-prefix sample trace variance falls to $0.83\times$ at
SFT, $0.37\times$ at DPO, and $0.43\times$ at RLVR; mean pairwise sample
distance rises slightly at SFT ($1.14\times$) before contracting to
$0.73\times$ and $0.76\times$ at DPO and RLVR. One caveat applies to the Base
reference point: Base continuations are shorter
(Table~\ref{tab:app-length}), which adds centroid-estimation noise. The sharp
contraction from SFT to DPO/RLVR, however, occurs between stages of comparable
realized length and is therefore not a length artifact. Since prefix, model
lineage, and decoding are all held fixed, this contraction shows post-training
narrowing the model's own continuation space and cannot be explained by
comparing one model against many human writers.

%% file: Sections/A07-1-prompt_interfeces_control.tex
\section{Prompt-Interface Control}
\label{app:prompt-interface-control}

The main stage-wise analysis compares an OLMo Base checkpoint that receives a
raw story prefix with instruction-tuned checkpoints that receive an explicit
continuation instruction through a chat-template interface. This means that the
Base-to-SFT contrast is not a pure training-stage causal effect: it combines a
change in model weights with a change in the input interface.

To estimate how much of the early-stage movement can be attributed to interface
framing alone, we run a prompt-interface control separately for the three story
domains. The control uses the same OLMo Base weights as the Raw Base condition,
but prepends the continuation instruction as plain text rather than applying a
chat template or adding special chat tokens. Thus the model is still used as a
base completion model, but the prompt contains the same task framing used for
the instruction-tuned continuations. All values are recomputed on matched
stories within each domain block and in each block's comparison space; they
should be interpreted within a domain rather than as replacements for the main
full-corpus estimates, and human rows should not be compared across blocks.

\begin{table*}[t]
\centering
\footnotesize
\setlength{\tabcolsep}{3.5pt}
\begin{tabular}{llrrrr}
\toprule
Domain & Source & Theme CV & Affective charge & Neutral & Style Var/Human \\
\midrule
\textit{New Yorker} & Human & 0.101 [0.100, 0.102] & 41.0 [40.5, 41.5] & 28.7 [28.3, 29.2] & 1.00 [0.93, 1.07] \\
 & Raw Base & 0.093 [0.092, 0.094] & 79.0 [78.5, 79.5] & 13.3 [13.0, 13.7] & 3.02 [2.99, 3.04] \\
 & Prompt-control Base & 0.100 [0.099, 0.100] & 68.9 [68.6, 69.2] & 18.8 [18.6, 19.0] & 0.70 [0.67, 0.73] \\
 & SFT & 0.086 [0.085, 0.086] & 36.5 [36.1, 36.9] & 35.0 [34.7, 35.3] & 0.52 [0.52, 0.53] \\
 & DPO & 0.079 [0.079, 0.079] & 22.0 [21.7, 22.2] & 43.1 [42.8, 43.4] & 0.34 [0.33, 0.35] \\
 & RLVR & 0.079 [0.079, 0.079] & 20.7 [20.4, 20.9] & 45.1 [44.8, 45.4] & 0.37 [0.36, 0.38] \\
\addlinespace
\textsc{TMAS} & Human & 0.095 [0.092, 0.098] & 25.4 [24.0, 26.8] & 35.7 [34.1, 37.3] & 1.00 [0.92, 1.08] \\
 & Raw Base & 0.093 [0.092, 0.095] & 44.6 [43.5, 45.6] & 28.5 [27.6, 29.4] & 0.75 [0.71, 0.78] \\
 & Prompt-control Base & 0.095 [0.093, 0.097] & 46.2 [45.2, 47.1] & 26.3 [25.5, 27.1] & 0.78 [0.74, 0.81] \\
 & SFT & 0.078 [0.077, 0.079] & 23.3 [22.4, 24.2] & 36.1 [35.1, 37.0] & 0.58 [0.54, 0.61] \\
 & DPO & 0.079 [0.078, 0.080] & 19.6 [18.8, 20.4] & 38.2 [37.1, 39.3] & 0.51 [0.47, 0.54] \\
 & RLVR & 0.079 [0.078, 0.080] & 18.7 [18.0, 19.5] & 39.4 [38.3, 40.4] & 0.52 [0.49, 0.56] \\
\addlinespace
\textsc{StoryStar} & Human & 0.107 [0.101, 0.116] & 22.6 [20.1, 25.1] & 38.4 [34.8, 41.8] & 1.00 [0.85, 1.14] \\
 & Raw Base & 0.109 [0.106, 0.114] & 37.1 [34.8, 39.6] & 32.0 [29.5, 34.4] & 0.95 [0.86, 1.04] \\
 & Prompt-control Base & 0.110 [0.106, 0.115] & 37.0 [34.6, 39.4] & 31.5 [29.1, 34.0] & 1.01 [0.90, 1.09] \\
 & SFT & 0.093 [0.090, 0.096] & 31.3 [28.2, 34.2] & 32.5 [30.2, 35.1] & 1.45 [1.19, 1.67] \\
 & DPO & 0.083 [0.082, 0.084] & 18.0 [16.8, 19.2] & 37.2 [35.4, 39.0] & 0.31 [0.27, 0.36] \\
 & RLVR & 0.083 [0.082, 0.084] & 16.9 [15.9, 18.1] & 38.7 [36.8, 40.5] & 0.33 [0.29, 0.38] \\
\bottomrule
\end{tabular}
\caption{
Prompt-interface control by story domain. Prompt-control Base uses the same
OLMo Base weights as Raw Base, but prepends the continuation instruction as a
plain-text prefix rather than using a chat template or special chat tokens.
Values are recomputed on matched stories within each domain block and in each
block's comparison space; comparisons should therefore be made within a domain
rather than across human rows, and these estimates should not be read as
replacements for the main full-corpus results.
}
\label{tab:app-prompt-interface-control}
\end{table*}

The control shows that prompt framing contributes to part of the
Base-to-instruction movement, but does not explain the main stage-wise
flattening pattern. The relevant comparison is within each domain block:
Prompt-control Base isolates the effect of adding instruction-like task framing
to the Base checkpoint, while SFT, DPO, and RLVR combine an instruction-facing
interface with post-trained weights.

\paragraph{Thematic rhythm.}
Thematic motion is the clearest case. Prompt-control Base stays close to Raw
Base in every domain, if anything moving slightly \emph{toward} the human value
rather than toward the post-trained endpoint:
$0.093 \rightarrow 0.100$ for \textit{The New Yorker},
$0.093 \rightarrow 0.095$ for \textsc{TMAS}, and
$0.109 \rightarrow 0.110$ for \textsc{StoryStar}. By contrast, the DPO/RLVR
endpoints are substantially lower in all three domains
($0.079$, $0.079$, and $0.083$). The post-training reduction in thematic CV
therefore cannot be attributed to the instruction-like prompt.

\paragraph{Affective prevalence.}
The affective facet is more prompt-sensitive, but in a way that is
structurally different from the post-training endpoint. In \textsc{TMAS} and
\textsc{StoryStar}, Prompt-control Base is essentially indistinguishable from
Raw Base in both affective charge and neutral prevalence. In \textit{The New
Yorker}, the instruction-like prompt does have a modest neutralizing effect,
lowering affective charge from $79.0\%$ to $68.9\%$ and raising neutral
narration from $13.3\%$ to $18.8\%$; but this remains far from the DPO/RLVR
endpoint (${\sim}21\%$ affective charge and ${\sim}45\%$ neutral). More
importantly, the underlying family decomposition shows that the prompt and the
post-training endpoint act through \emph{different} mechanisms. Prompt-control
Base lowers conflict while \emph{raising} surprise--curiosity, redistributing
affect across families rather than suppressing it overall. The DPO/RLVR
endpoint does the opposite: both conflict and surprise--curiosity are
suppressed relative to Human while neutral rises. Affective flattening at the
post-trained endpoints is therefore an overall suppression of high-arousal
affect, not the within-affect redistribution induced by prompt framing.

\paragraph{Linguistic diversity.}
Style variance is the most interface-sensitive facet, and we treat it with the
most caution. For \textit{The New Yorker}, instruction-like prompting alone
sharply narrows Base outputs ($3.02 \rightarrow 0.70$ Var/Human), so the
Raw Base-to-SFT style contrast in this domain should be read as a mixture of
interface and training effects rather than a pure training effect. We do not
claim otherwise. Two observations nonetheless show that the style result is not
merely a prompt artifact. First, this large interface-driven narrowing does not
appear in \textsc{TMAS} or \textsc{StoryStar}, where Prompt-control Base remains
close to Raw Base ($0.75 \rightarrow 0.78$ and $0.95 \rightarrow 1.01$). Second,
even in \textit{The New Yorker}, post-training narrows style \emph{further}
beyond the prompt-induced level: the DPO and RLVR endpoints reach
$0.34$--$0.37$ Var/Human, well below the $0.70$ produced by prompting alone.
The later post-training stages thus add compression on top of any interface
effect.

\paragraph{Summary.}
Taken together, the control rules out a broad prompt-artifact explanation of
narrative flattening: thematic compression is unaffected by the interface,
affective compression operates through a different mechanism than prompt
framing, and stylistic compression continues to accumulate after the
prompt-induced narrowing. We retain one specific caution---the Raw Base-to-SFT
style contrast for \textit{The New Yorker} mixes interface and training
effects---and note that the SFT-to-DPO-to-RLVR comparisons, which share the
instruction interface throughout, more directly isolate successive
post-training stages.

%% file: Sections/A08-Full-Cross-Domain-Result.tex
\section{Full Results and Uncertainty Estimates}
\label{app:full-results}

This appendix consolidates the per-domain, per-stage results and their
uncertainty estimates. Unless otherwise noted, confidence intervals are 95\%
bootstrap intervals computed by resampling stories. The exception is
StyleDistance MMD: for human--model MMD curves, confidence intervals use the
sentence-state bootstrap described in Appendix~\ref{app:embedding-details}.
These intervals quantify uncertainty in the sampled sentence-state distribution
rather than story-block uncertainty. Across-story style variance uses
story-level bootstrap over continuation centroids. Affective-charge intervals
are conservative component-bound intervals obtained by summing the lower and
upper bounds for surprise--curiosity and conflict.

\subsection{Per-Domain, Per-Stage Results}
\label{app:full-cross-corpus}

Table~\ref{tab:app-full-results} reports the full results for the three main
facets across all domains and training stages. These values complement the
main-text figures, which emphasize the \textit{New Yorker} stage-wise
trajectory and the Human-vs-RLVR cross-domain endpoint comparison.

\begin{table*}[t]
\centering
\footnotesize
\setlength{\tabcolsep}{2pt}
\begin{tabular}{lrrrrrr}
\toprule
\textbf{Source} &
\textbf{Theme CV} &
\textbf{Affect. charge} &
\textbf{Conflict} &
\textbf{Surprise--cur.} &
\textbf{Neutral} &
\textbf{Style MMD$^2$} \\
\midrule
\multicolumn{7}{l}{\textit{New Yorker}} \\
Human & 0.104 [0.104, 0.105] & 41.0 [40.2, 41.7] & 20.0 [19.6, 20.3] & 21.0 [20.6, 21.4] & 28.7 [28.3, 29.2] & --- \\
Base & 0.096 [0.096, 0.097] & 79.2 [77.3, 81.0] & 46.6 [45.5, 47.8] & 32.5 [31.8, 33.2] & 13.2 [12.9, 13.6] & 0.245 [0.226, 0.268] \\
SFT & 0.089 [0.088, 0.089] & 36.6 [36.1, 37.1] & 10.6 [10.5, 10.8] & 26.0 [25.6, 26.3] & 35.0 [34.7, 35.2] & 0.406 [0.378, 0.432] \\
DPO & 0.081 [0.081, 0.081] & 22.0 [21.6, 22.3] & 8.4 [8.2, 8.5] & 13.6 [13.4, 13.7] & 43.1 [42.8, 43.4] & 0.532 [0.499, 0.563] \\
RLVR & 0.081 [0.081, 0.081] & 20.7 [20.3, 21.0] & 7.7 [7.5, 7.8] & 13.0 [12.8, 13.1] & 45.1 [44.7, 45.4] & 0.516 [0.486, 0.548] \\
\midrule
\multicolumn{7}{l}{\textsc{TMAS}} \\
Human & 0.098 [0.095, 0.101] & 26.5 [24.6, 28.3] & 11.2 [10.3, 12.1] & 15.2 [14.3, 16.2] & 36.2 [34.7, 37.9] & --- \\
Base & 0.098 [0.097, 0.100] & 44.9 [43.5, 46.4] & 12.2 [11.7, 12.8] & 32.7 [31.8, 33.6] & 28.8 [27.9, 29.7] & 0.088 \\
SFT & 0.080 [0.079, 0.081] & 24.3 [23.0, 25.5] & 8.2 [7.6, 8.8] & 16.1 [15.4, 16.7] & 36.6 [35.7, 37.6] & 0.054 \\
DPO & 0.081 [0.080, 0.082] & 20.3 [19.2, 21.3] & 6.4 [5.9, 6.9] & 13.9 [13.4, 14.5] & 39.0 [37.9, 40.2] & 0.025 \\
RLVR & 0.081 [0.081, 0.082] & 19.5 [18.4, 20.6] & 5.9 [5.4, 6.5] & 13.6 [13.0, 14.2] & 40.2 [39.0, 41.2] & 0.018 \\
\midrule
\multicolumn{7}{l}{\textsc{StoryStar}} \\
Human & 0.110 [0.103, 0.118] & 23.2 [20.3, 26.4] & 9.9 [8.5, 11.5] & 13.3 [11.9, 14.9] & 38.6 [35.4, 41.8] & --- \\
Base & 0.113 [0.109, 0.118] & 33.4 [30.5, 36.5] & 9.7 [8.5, 11.2] & 23.7 [22.0, 25.4] & 34.3 [31.7, 37.1] & 0.024 \\
SFT & 0.095 [0.092, 0.098] & 25.5 [23.2, 28.0] & 8.4 [7.5, 9.5] & 17.1 [15.7, 18.5] & 36.1 [34.2, 38.0] & 0.021 \\
DPO & 0.085 [0.083, 0.086] & 18.4 [16.8, 20.0] & 5.8 [5.1, 6.6] & 12.6 [11.8, 13.4] & 37.2 [35.4, 38.8] & 0.024 \\
RLVR & 0.085 [0.084, 0.086] & 17.3 [15.7, 19.0] & 5.2 [4.6, 5.9] & 12.1 [11.2, 13.1] & 38.8 [37.0, 40.7] & 0.018 \\
\bottomrule
\end{tabular}
\caption{
Full per-domain, per-stage results with 95\% bootstrap confidence intervals;
each domain forms one block. Theme CV is the coefficient of variation of
adjacent-sentence topic-jump $L_2$ distances. Affective values are percentages
of sentences assigned to each affect family; \emph{Affect.\ charge} is
surprise--curiosity plus conflict, and \emph{Surprise--cur.} abbreviates the
surprise--curiosity family.
Style MMD$^2$ is measured to the matched human reference for each domain;
cross-corpus style MMD intervals for \textsc{TMAS} and \textsc{StoryStar}
were not stored, so point estimates are reported for those rows.
}
\label{tab:app-full-results}
\end{table*}

\subsection{Supplementary Stage-Wise Statistics}
\label{app:supplementary-stage}

Tables~\ref{tab:app-topic-stage-ci} and~\ref{tab:app-style-progression-ci}
report statistics not captured in the full results table above: the
distributional spread of thematic unevenness across stories, and the
across-story style variance at each training stage.

The thematic drop is monotonic from Human through SFT and then saturates at
DPO/RLVR, with the 5--95\% range narrowing substantially
(Table~\ref{tab:app-topic-stage-ci}). Post-training simultaneously increases
the distance from the human style-neural reference and sharply reduces
across-story style variation (Table~\ref{tab:app-style-progression-ci}).

\begin{table*}
\centering
\small
\begin{tabular}{lrrr}
\toprule
Stage & Mean CV [95\% CI] & 5--95\% range & Loss vs.\ human \\
\midrule
Human & 0.104 [0.104, 0.105] & [0.063, 0.162] & 0.0\% \\
Base  & 0.096 [0.096, 0.097] & [0.027, 0.188] & 8.0\% \\
SFT   & 0.089 [0.088, 0.089] & [0.052, 0.146] & 15.1\% \\
DPO   & 0.081 [0.081, 0.081] & [0.056, 0.108] & 22.2\% \\
RLVR  & 0.081 [0.081, 0.081] & [0.056, 0.108] & 22.2\% \\
\bottomrule
\end{tabular}
\caption{
Stage-wise thematic unevenness (\textit{New Yorker} continuations).
The 5--95\% range matches the distribution panel in
Figure~\ref{fig:stage-progression-topic}B.
}
\label{tab:app-topic-stage-ci}
\end{table*}

\begin{table*}[t]
\centering
\small
\begin{tabular}{lcc}
\toprule
Stage & MMD\textsuperscript{2} to human & Var/human \\
\midrule
Human & --- (ref) & 1.000 [0.932, 1.062] \\
Base  & 0.245 [0.226, 0.268] & 6.026 [5.963, 6.079] \\
SFT   & 0.406 [0.378, 0.432] & 0.835 [0.818, 0.856] \\
DPO   & 0.532 [0.499, 0.563] & 0.485 [0.473, 0.497] \\
RLVR  & 0.516 [0.486, 0.548] & 0.521 [0.508, 0.534] \\
\bottomrule
\end{tabular}
\caption{
Style divergence and across-story variance for \textit{New Yorker}
continuations. Var/human normalizes across-story style variance by the
human baseline. MMD\textsuperscript{2} intervals use the sentence-state
bootstrap; Var/human intervals use story-level bootstrap.
}
\label{tab:app-style-progression-ci}
\end{table*}

\subsection{Endpoint Convergence Across Domains}
\label{app:endpoint-convergence}

Table~\ref{tab:app-endpoint-range-ci} reports confidence intervals for the
cross-domain range reductions in Figure~\ref{fig:convergence}. For each
bootstrap replicate, stories are resampled within each domain and endpoint,
domain means are recomputed, and the cross-domain range is measured as the
maximum domain mean minus the minimum.

\begin{table*}[t]
\centering
\small
\begin{tabular}{lccc}
\toprule
Metric & Human range & RLVR range & Range reduction \\
\midrule
Topical jump CV &
0.0116 [0.0050, 0.0205] &
0.0037 [0.0024, 0.0052] &
62.1\% \\
Affective charge pp. &
17.8 [15.2, 20.2] &
3.3 [2.1, 4.5] &
81.3\% \\
\bottomrule
\end{tabular}
\caption{
Cross-domain endpoint convergence with 95\% bootstrap confidence intervals.
Range reduction is computed from the point estimates in the first two columns.
}
\label{tab:app-endpoint-range-ci}
\end{table*}

For the style panel in Figure~\ref{fig:convergence}, the visualization is a
PCA projection of style-neural embeddings rather than a scalar estimator. The
corresponding scalar evidence is the cross-domain style MMD comparison reported
in the main text: human style baselines are far apart, especially
\textit{New Yorker} versus the two common-story corpora, whereas the three RLVR
endpoints lie close to one another in the same style-neural space.

%% file: Sections/A09-Qwen.tex
\section{Additional Model-Family Endpoint Checks}
\label{app:additional-models}

To test whether the observed direction is specific to the OLMo instruction
path, we repeat the matched-continuation analysis using Base-vs-Instruct
endpoint comparisons from three additional model families: Qwen2.5-32B,
Llama-3.1-8B, and Gemma-3-12B. These comparisons are not stage-wise: each contrasts a base
checkpoint with an instruction-tuned endpoint. They therefore test directional
cross-family robustness, but do not identify which post-training stage
contributes to compression.
Human rows are recomputed on the exact matched subset available for each
model-family/domain block. Endpoint comparisons should therefore be interpreted
within each block rather than by comparing human baselines across model
families.

\begin{table*}[t]
\centering
\footnotesize
\setlength{\tabcolsep}{3.5pt}
\begin{tabular}{llrrrrr}
\toprule
\textbf{Domain} &
\textbf{Source} &
\textbf{Theme CV} &
\textbf{Affect. charge} &
\textbf{Neutral} &
\textbf{MMD$^2$} &
\textbf{Var/Human} \\
\midrule
\multicolumn{7}{l}{\textit{Qwen2.5-32B}} \\
New Yorker & Human & 0.104 [0.103, 0.106] & 41.0 [40.5, 41.4] & 28.7 [28.3, 29.2] & --- & 1.00 [0.93, 1.07] \\
 & Base & 0.093 [0.092, 0.093] & 67.6 [67.3, 67.9] & 19.1 [18.9, 19.4] & 0.931 & 0.41 [0.39, 0.44] \\
 & Instruct & 0.086 [0.085, 0.086] & 38.9 [38.6, 39.2] & 29.5 [29.2, 29.7] & 1.570 & 0.30 [0.30, 0.31] \\
\addlinespace
TMAS & Human & 0.095 [0.093, 0.099] & 25.4 [24.0, 26.7] & 35.7 [34.3, 37.5] & --- & 1.00 [0.91, 1.09] \\
 & Base & 0.089 [0.088, 0.091] & 43.8 [43.0, 44.7] & 28.5 [27.6, 29.4] & 0.871 & 0.42 [0.38, 0.45] \\
 & Instruct & 0.085 [0.083, 0.086] & 29.1 [28.1, 30.0] & 28.7 [27.8, 29.7] & 0.582 & 0.51 [0.46, 0.57] \\
\addlinespace
StoryStar & Human & 0.107 [0.101, 0.116] & 22.6 [20.0, 25.0] & 38.4 [34.9, 41.8] & --- & 1.00 [0.82, 1.19] \\
 & Base & 0.106 [0.101, 0.113] & 30.0 [27.9, 32.3] & 35.3 [32.4, 38.0] & 0.040 & 0.82 [0.68, 0.96] \\
 & Instruct & 0.094 [0.092, 0.097] & 28.6 [26.8, 30.4] & 39.9 [37.7, 42.0] & 0.201 & 0.42 [0.35, 0.49] \\
\midrule
\multicolumn{7}{l}{\textit{Llama-3.1-8B}} \\
New Yorker & Human & 0.103 [0.101, 0.104] & 37.2 [36.4, 38.0] & 31.4 [30.6, 32.1] & --- & 1.00 [0.95, 1.04] \\
 & Base & 0.084 [0.082, 0.085] & 68.4 [67.7, 69.0] & 22.8 [22.3, 23.3] & 0.282 & 0.39 [0.38, 0.41] \\
 & Instruct & 0.068 [0.068, 0.069] & 35.3 [34.1, 36.5] & 37.6 [36.9, 38.3] & 0.578 & 0.41 [0.39, 0.42] \\
\addlinespace
TMAS & Human & 0.095 [0.093, 0.099] & 25.4 [24.0, 26.7] & 35.7 [34.3, 37.5] & --- & 1.00 [0.91, 1.09] \\
 & Base & 0.101 [0.099, 0.103] & 58.4 [57.1, 59.7] & 23.9 [23.1, 24.6] & 0.995 & 0.75 [0.69, 0.81] \\
 & Instruct & 0.081 [0.079, 0.083] & 55.9 [53.6, 58.0] & 27.0 [26.3, 27.7] & 0.905 & 1.35 [1.21, 1.49] \\
\addlinespace
StoryStar & Human & 0.107 [0.101, 0.116] & 22.6 [20.0, 25.0] & 38.4 [34.9, 41.8] & --- & 1.00 [0.83, 1.16] \\
 & Base & 0.102 [0.098, 0.106] & 57.2 [54.9, 59.5] & 26.0 [24.6, 27.4] & 0.802 & 0.54 [0.46, 0.62] \\
 & Instruct & 0.082 [0.080, 0.085] & 55.8 [52.2, 59.4] & 38.3 [26.6, 40.0] & 0.946 & 0.66 [0.53, 0.77] \\
\midrule
\multicolumn{7}{l}{\textit{Gemma-3-12B}} \\
New Yorker & Human & 0.107 [0.104, 0.110] & 33.1 [32.0, 34.1] & 34.1 [33.0, 35.3] & --- & 1.00 [0.96, 1.03] \\
 & Base & 0.096 [0.094, 0.097] & 55.1 [54.0, 56.0] & 24.0 [23.4, 24.7] & 0.310 & 0.47 [0.45, 0.49] \\
 & Instruct & 0.095 [0.094, 0.096] & 32.8 [32.0, 33.6] & 36.8 [36.1, 37.4] & 0.437 & 0.09 [0.08, 0.10] \\
\addlinespace
TMAS & Human & 0.095 [0.093, 0.099] & 25.4 [24.0, 26.7] & 35.7 [34.3, 37.5] & --- & 1.00 [0.91, 1.09] \\
 & Base & 0.084 [0.083, 0.086] & 48.0 [46.3, 49.7] & 25.7 [24.6, 26.8] & 0.702 & 0.67 [0.62, 0.72] \\
 & Instruct & 0.080 [0.079, 0.081] & 27.9 [26.9, 28.9] & 34.4 [33.4, 35.4] & 0.460 & 0.29 [0.26, 0.32] \\
\addlinespace
StoryStar & Human & 0.107 [0.101, 0.116] & 22.6 [20.0, 25.0] & 38.4 [34.9, 41.8] & --- & 1.00 [0.82, 1.18] \\
 & Base & 0.097 [0.091, 0.105] & 46.9 [44.1, 49.7] & 27.3 [24.8, 29.9] & 0.467 & 0.63 [0.53, 0.72] \\
 & Instruct & 0.096 [0.094, 0.098] & 24.9 [23.3, 26.4] & 34.5 [32.4, 36.5] & 0.163 & 0.28 [0.23, 0.35] \\
\bottomrule
\end{tabular}
\caption{
Additional Base-vs-Instruct endpoint checks across model families and story
domains. Values are mean [95\% CI] where available. Theme CV is the coefficient
of variation of adjacent-sentence topic-jump $L_2$ distances. \emph{Affect.
charge} is surprise--curiosity plus conflict, in percentage points.
\emph{MMD$^2$} is the style-neural distance to the matched human reference for
each domain, and \emph{Var/Human} normalizes across-story style variance by the
matched human baseline.
These endpoint comparisons test facet-level Base-to-Instruct robustness across
model families. They do not provide stage-wise attribution and should not be
read as full replications of the OLMo post-training trajectory.
}
\label{tab:app-additional-models}
\end{table*}

\paragraph{Reading the endpoint checks.}
These comparisons are endpoint checks rather than stage-wise replications.
Each compares a base checkpoint with an instruction-tuned endpoint from the same
model family. They therefore test whether the direction of Base-to-Instruct
movement generalizes beyond OLMo, but they do not identify which post-training
stage produces the movement.

\paragraph{Thematic motion.}
All three additional model families show Base-to-Instruct reductions in
thematic CV across the three domains, although the magnitude varies by family.
For Qwen2.5-32B, thematic CV falls from $0.093 \to 0.086$ on
\textit{The New Yorker}, $0.089 \to 0.085$ on \textsc{TMAS}, and
$0.106 \to 0.094$ on \textsc{StoryStar}. For Llama-3.1-8B, the same
direction holds more strongly: $0.084 \to 0.068$, $0.101 \to 0.081$,
and $0.102 \to 0.082$. Gemma-3-12B also moves in the same direction,
with smaller reductions on \textit{The New Yorker} and \textsc{StoryStar}
($0.096 \to 0.095$ and $0.097 \to 0.096$) and a clearer reduction on
\textsc{TMAS} ($0.084 \to 0.080$). These endpoint checks support the
claim that thematic regularization is not unique to the OLMo lineage, while
also showing that its magnitude is family-dependent.

\paragraph{Affective prevalence.}
Affective compression is the most consistent cross-family signal. Across all
nine family--domain endpoint comparisons, instruction tuning lowers affective
charge relative to the corresponding base model. The New Yorker drop is large
in every family: $67.6\% \to 38.9\%$ for Qwen2.5-32B,
$68.4\% \to 35.3\%$ for Llama-3.1-8B, and $55.1\% \to 32.8\%$ for
Gemma-3-12B. Gemma shows the same direction on \textsc{TMAS}
($48.0\% \to 27.9\%$) and \textsc{StoryStar} ($46.9\% \to 24.9\%$).
Neutral narration also rises in nearly all settings, including all Gemma and
Llama domains. These results make affective neutralization the most robust
facet across model families.

\paragraph{Linguistics Diversity.}
The style results are the most domain- and family-dependent. Qwen2.5-32B
shows lower across-story style variance at the instruct endpoint for
\textit{The New Yorker} and \textsc{StoryStar}, but not for \textsc{TMAS}.
Llama-3.1-8B does not show uniform Base-to-Instruct variance compression:
style variance rises in the two common-fiction domains. Gemma-3-12B, by
contrast, shows strong variance compression in all three domains
($0.47 \to 0.09$ for \textit{The New Yorker}, $0.67 \to 0.29$ for
\textsc{TMAS}, and $0.63 \to 0.28$ for \textsc{StoryStar}). Its MMD pattern,
however, is domain-sensitive: distance to the New Yorker human style reference
increases, while distance to the two common-fiction references decreases. This
is consistent with a narrow instruct-style attractor that sits closer to
common-fiction baselines than to professional literary fiction. Overall, style
compression is less uniformly stage-general than affective neutralization, but
the professional-fiction distortion is supported across endpoint checks.

\paragraph{Takeaway.}
The additional endpoint checks therefore support a narrower and more reliable
generalization: thematic regularization and affective neutralization appear
across model families, while style compression is strongest for professional
fiction and more variable for common-fiction domains. The coordinated
three-facet, stage-wise compression reported in the main text remains clearest
in the OLMo instruction path, where intermediate checkpoints allow us to trace
how compression accumulates across post-training stages.

%% file: Sections/A10-Cut-point-Robustness.tex
\section{Cut-Point Robustness}
\label{app:cutpoint}

To test whether narrative flattening is specific to early, middle, late, or
near-ending continuation contexts, we repeat the main analyses separately for
prefix cut points at 40\%, 60\%, 80\%, and 90\% of each story's sentence
sequence. Table~\ref{tab:app-cutpoint} reports the cut-point-specific values
for the \textit{New Yorker} OLMo instruction path.

\begin{table*}[t]
\centering
\footnotesize
\setlength{\tabcolsep}{4pt}
\begin{tabular}{llrrrr}
\toprule
Cut & Stage & Theme CV & Affective charge & Neutral & Style MMD$^2$ \\
\midrule
40\% & Human & 0.107 [0.106, 0.108] & 42.0 [41.3, 42.6] & 29.6 [29.2, 30.0] & --- \\
 & Base & 0.106 [0.105, 0.107] & 70.7 [69.1, 72.3] & 19.0 [18.5, 19.4] & 0.296 [0.272, 0.320] \\
 & SFT & 0.098 [0.097, 0.099] & 42.5 [41.8, 43.2] & 32.5 [32.2, 32.9] & 0.413 [0.389, 0.440] \\
 & DPO & 0.082 [0.082, 0.083] & 22.6 [22.2, 22.9] & 43.3 [43.0, 43.6] & 0.515 [0.476, 0.543] \\
 & RLVR & 0.083 [0.082, 0.083] & 21.1 [20.8, 21.4] & 45.6 [45.3, 46.0] & 0.524 [0.492, 0.555] \\
\midrule
60\% & Human & 0.105 [0.104, 0.106] & 41.8 [41.0, 42.5] & 29.0 [28.6, 29.5] & --- \\
 & Base & 0.103 [0.101, 0.104] & 77.9 [75.7, 80.2] & 14.0 [13.6, 14.5] & 0.270 [0.249, 0.292] \\
 & SFT & 0.094 [0.093, 0.094] & 40.2 [39.5, 40.9] & 33.1 [32.8, 33.4] & 0.393 [0.363, 0.421] \\
 & DPO & 0.083 [0.082, 0.083] & 22.7 [22.3, 23.0] & 42.5 [42.2, 42.9] & 0.486 [0.457, 0.521] \\
 & RLVR & 0.083 [0.082, 0.083] & 21.3 [21.0, 21.7] & 44.5 [44.2, 44.9] & 0.513 [0.483, 0.541] \\
\midrule
80\% & Human & 0.103 [0.102, 0.105] & 40.6 [39.8, 41.5] & 28.4 [27.9, 28.9] & --- \\
 & Base & 0.086 [0.084, 0.087] & 83.2 [80.6, 85.6] & 10.6 [10.1, 11.0] & 0.220 [0.197, 0.240] \\
 & SFT & 0.084 [0.084, 0.085] & 34.3 [33.7, 34.9] & 35.8 [35.5, 36.3] & 0.432 [0.406, 0.462] \\
 & DPO & 0.081 [0.080, 0.081] & 22.0 [21.6, 22.4] & 42.8 [42.4, 43.1] & 0.537 [0.504, 0.569] \\
 & RLVR & 0.081 [0.081, 0.081] & 20.7 [20.3, 21.1] & 44.7 [44.3, 45.1] & 0.503 [0.471, 0.541] \\
\midrule
90\% & Human & 0.100 [0.099, 0.102] & 39.5 [38.3, 40.5] & 27.9 [27.3, 28.6] & --- \\
 & Base & 0.075 [0.073, 0.078] & 84.9 [82.7, 87.2] & 9.3 [8.8, 9.8] & 0.168 [0.152, 0.189] \\
 & SFT & 0.078 [0.078, 0.079] & 29.2 [28.4, 29.8] & 38.4 [37.9, 38.8] & 0.435 [0.406, 0.463] \\
 & DPO & 0.078 [0.078, 0.079] & 20.6 [20.1, 21.1] & 43.8 [43.4, 44.2] & 0.524 [0.491, 0.564] \\
 & RLVR & 0.078 [0.077, 0.078] & 19.5 [19.0, 20.0] & 45.4 [44.9, 45.8] & 0.509 [0.481, 0.540] \\
\bottomrule
\end{tabular}
\caption{
Cut-point robustness for \textit{New Yorker} continuations. Theme CV is the
per-story coefficient of variation of adjacent-sentence topic-jump $L_2$
distances. Affective charge is surprise--curiosity plus conflict prevalence
in percentage points; its interval is a conservative component-bound interval
obtained by summing the lower and upper bounds for the two component families.
Neutral is reported in percentage points. Style MMD$^2$ is measured to the
matched human style-neural reference at the same cut point. Values are mean
[95\% CI].
}
\label{tab:app-cutpoint}
\end{table*}

Across cut points, the direction of compression is stable. Thematic CV is
lower at the DPO/RLVR endpoints than in human continuations at every cut point;
affective charge falls sharply after SFT; and neutral narration rises above
the human baseline after instruction-stage post-training. Near-ending
continuations show smaller absolute room for variation because suffixes are
shorter, but the qualitative direction is unchanged.

%% file: Sections/A12-Ethics-Statement.tex
\section{Data Release and Reproducibility}
\label{app:ethics}

\paragraph{Copyrighted literary text.}
Some source texts are copyrighted. We use them only for non-commercial,
non-distributing research and report only aggregate statistics. We do not release
raw New Yorker stories, long excerpts, or generated continuations that contain
substantial copyrighted context. Any released code will operate on user-provided
texts or on public-domain examples.

\paragraph{Generated continuations.}
Generated continuations are used to compute aggregate narrative metrics. Because
the continuations are conditioned on copyrighted prefixes in some domains, we do
not release the full generated text for those domains. We may release aggregate
metric tables, plotting scripts, and anonymized metadata that do not contain
protected story text.

\paragraph{Human and LLM-assisted affect labels.}
The affect adaptation set uses external literary prose and a hybrid annotation
process. We release annotation guidelines and aggregate validation statistics,
but we do not release copyrighted sentence text from restricted sources. If a
public-domain-only subset is used, we release that subset when licensing permits.

\paragraph{Privacy and author identifiers.}
For public-platform data, we remove author names and other obvious identifiers
before analysis. Our unit of analysis is the story text and its aggregate
continuation metrics, not individual authors.

\paragraph{Reproducibility.}
We release an anonymized software package for preprocessing, continuation segmentation, metric computation, statistical analysis, and figure generation.
The package does not include restricted source texts, raw \textit{New Yorker} stories, long excerpts, or full generated continuations conditioned on copyrighted prefixes. 
Instead, the code operates on user-provided corpora and included public-domain or synthetic examples, and we provide aggregate metric tables sufficient to reproduce the reported figures and summary statistics where redistribution of underlying text is not permitted. Exact reproduction of stochastic generations may differ because no explicit decoding seed was passed; however, all reported analyses are computed from archived generations, and downstream metric computation is deterministic.

%% file: Sections/A11-Qualitative-Illustration.tex
\section{Qualitative Illustrations}
\label{app:qualitative}

This appendix offers a short qualitative illustration of the distributional
effects quantified in the main text. It is \emph{not} a human-validation study:
the excerpts below are illustrative only and are not used as evidence for any
claim. All quantitative conclusions rest on the metrics and statistical tests
reported in Section~\ref{sec:results} and the preceding appendices. The metric
values in the small tables are computed over the full continuation, not over the
displayed snippet.

Consistent with the copyright constraints in Appendix~\ref{app:ethics}, we avoid
\textit{The New Yorker} excerpts and show only brief snippets from non-restricted
sources, paraphrasing where redistribution rights are unclear. Examples were
selected to illustrate the aggregate direction of the metrics, not to estimate
effect frequency; the frequency and magnitude of the effects are reported in the
full quantitative tables.

For each example we pair a shared human-written prefix with the matched human
continuation, the base-model continuation, and the RLVR continuation. The
examples make the measured contrasts concrete at the level of text: more
uniformly sized thematic motion, muted affect, and a narrower stylistic
register.

\paragraph{\textsc{Tell Me A Story} (prompt-guided fiction).}
\textit{Shared prefix summary.} A painter wakes after an uncanny episode in
which a painted room seemed to cross into ordinary life.
\begin{itemize}[noitemsep, topsep=2pt, leftmargin=*]
  \item \textbf{Human continuation.} ``He smelled bacon frying in the kitchen
  and knew Paul must be cooking breakfast.''
  \item \textbf{Base continuation.} ``Something was terribly wrong. Paul. The
  painting. There was this paralyzing trepidation.''
  \item \textbf{RLVR continuation.} ``He sat up slowly, joints creaking, but
  felt strangely refreshed.''
\end{itemize}

\begin{center}
\small
\begin{tabular}{lrrr}
\toprule
Continuation & Theme CV & Affective charge & Neutral \\
\midrule
Human & 0.131 & 49.4\% & 32.2\% \\
Base & 0.107 & 41.7\% & 50.0\% \\
RLVR & 0.085 & 17.0\% & 67.9\% \\
\bottomrule
\end{tabular}
\end{center}

In this example, the human continuation begins with ordinary domestic action
but keeps the uncanny premise active. The base continuation is more overtly
marked by threat and tension. The RLVR continuation is smoother and more neutral
in register; over the full continuation, theme CV and affective charge fall
while neutral narration increases.

\paragraph{\textsc{StoryStar} (public-platform fiction).}
\textit{Shared prefix summary.} Two non-human atmospheric beings debate whether
intervening in a human crisis would also serve their own survival.
\begin{itemize}[noitemsep, topsep=2pt, leftmargin=*]
  \item \textbf{Human continuation.} ``First we save an intelligent species
  from extinction.''
  \item \textbf{Base continuation.} ``Felp shrank his pressure membrane into a
  ball and sent billions of identical messages.''
  \item \textbf{RLVR continuation.} ``We save ourselves, and them.''
\end{itemize}

\begin{center}
\small
\begin{tabular}{lrrr}
\toprule
Continuation & Theme CV & Affective charge & Neutral \\
\midrule
Human & 0.142 & 42.6\% & 48.9\% \\
Base & 0.136 & 45.1\% & 47.1\% \\
RLVR & 0.071 & 11.1\% & 66.7\% \\
\bottomrule
\end{tabular}
\end{center}

Here the base-model sample remains closer to the human continuation's tension
and decision structure, while the RLVR sample compresses the decision into a
cleaner moral formulation. The example mirrors the aggregate pattern: lower
thematic unevenness, lower affective charge, and higher neutral narration.

\subsection{What a Thematic-CV Gap Looks Like: Splice vs.\ Bridge}
\label{app:splice-vs-bridge}

The quantitative case that a lower CV means flatter rhythm rather than tighter
local cohesion is made in Appendix~\ref{app:metric-validity}. This section shows
what that difference looks like in text. Reading matched Human-vs-RLVR pairs
with large CV gaps, the contrast reduces to a single recurring mechanism, which
we call \emph{splice vs.\ bridge}: both continuations reach the same narrative
beats, but the human makes unbridged cuts where the aligned model lays down
connective tissue that evens out the rhythm. Each pair below continues an
identical human prefix; CV values are computed over the full continuation, not
the displayed snippet. Following Appendix~\ref{app:ethics} we draw only on
non-restricted sources.

\paragraph{(a) A same-destination natural experiment} (human CV $0.23$ vs.\
RLVR $0.14$). Two characters realize they belonged to the same college club;
both continuations deliver that recognition, so content is held fixed and only
rhythm differs.
\begin{itemize}[noitemsep, topsep=2pt, leftmargin=*]
  \item \textbf{Human.} ``\textit{Rosebud Flightbloom! You were a legend in the
  blind river games!}'' ``\textit{No! No way! You were in the Cult of the Purple
  Dinosaur Flower? Really?}'' Rosalie exclaimed \ldots{} Dr.\ Diaz \ldots{}
  pulled out their own lumpy, shiny purple stone. Rosalie reached into her
  borrowed blazer and produced a nearly identical stone. Both burst out
  laughing.
  \item \textbf{RLVR.} ``\textit{You led the artifact retrieval team. I followed
  your group's dispatch reports \ldots{} You saved that rare seed from
  floodwaters---and your account of the cave-in was cited in my thesis. I've
  admired your work ever since.}'' Rosalie's jaw dropped open \ldots{}
\end{itemize}
The human oscillates across kinds of beat---exclamatory dialogue, a physical
action (producing the stones), shared laughter---in quick, unequal jumps; the
model delivers one continuous, evenly paced admiring speech.

\paragraph{(b) Tonal interruption} (human CV $0.42$ vs.\ RLVR $0.11$).
\begin{itemize}[noitemsep, topsep=2pt, leftmargin=*]
  \item \textbf{Human.} It is nice to just literally float and be moved about by
  the gentle waves. I wish this moment could last forever. Clang! Clang! Ugh! It
  is the bell from the boat calling us to return for lunch \ldots{} My
  snorkeling time is sadly over. However, I'll always remember my time exploring
  the sea.
  \item \textbf{RLVR.} But just beneath the pleasant quiet, a faint, sinuous
  shadow undulates along the sand. I freeze. My heart skips \ldots{} instinct
  takes over and I tuck my knees to my chest and hover just above the bottom.
  The movement stirs up the silt and the shadow wavers and dissolves.
\end{itemize}
The human dwells (stillness, a wish), is ruptured by an offstage sound, cuts to
departure, and closes on reflection---three registers in a few sentences. The
model sustains a single, continuously escalating suspense beat.

\paragraph{(c) Unbridged time-skip} (human CV $0.25$ vs.\ RLVR $0.14$).
\begin{itemize}[noitemsep, topsep=2pt, leftmargin=*]
  \item \textbf{Human.} \ldots{} for tonight, we are having another round
  together \ldots{} We start to sing together, ``\textit{we're not going to bake
  it!}'' \ldots{} Don't worry, we're all taking cabs home. Election day is
  finally here, my friends. All of us are at the pub with Sam.
  \item \textbf{RLVR.} As I twist off the cap of my beer, a new idea bubbles up
  \ldots{} ``\textit{We need music. Everyone remembers Footloose, right?}''
  \ldots{} plans start bubbling fast as froth on our beers: a street dance
  \ldots{} a dueling stereo system. Flashing lights.
\end{itemize}
The human jumps from ``tonight'' to ``Election day is finally here'' with no
transition sentence; the model stays inside one continuous brainstorm. Two
further splice types recur and are omitted for space: register or frame breaks,
and syntactic fracture (e.g., ``I could not. Think.'').

\paragraph{Reading.}
In each pair the model continuation is the more locally coherent one, and that
smoothness is precisely what the low CV records. The human's higher CV comes
from deliberate discontinuity---unbridged time-skips, interruptions, register
breaks---not from incoherence. A lower CV therefore indicates a flatter, more
regularized rhythm rather than merely tighter local cohesion, which is the same
conclusion the shuffle null and the numerator/denominator decomposition reach
quantitatively (Appendices~\ref{app:shuffle-null}
and~\ref{app:cv-decomposition}).